\documentclass[lettersize,journal]{IEEEtran}
\def\ARXIVMAINDOC{1}
\usepackage{amsmath,amsfonts}
\usepackage{algpseudocode}
\usepackage{algorithm}
\usepackage{array}
\usepackage[caption=false,font=normalsize,labelfont=sf,textfont=sf]{subfig}
\usepackage{textcomp}
\usepackage{stfloats}
\usepackage{url}
\usepackage{verbatim}
\usepackage{graphicx}
\usepackage{cite}
\usepackage[table]{xcolor}
\usepackage{booktabs}
\usepackage{pifont}
\usepackage{amssymb}
\usepackage{multirow}
\usepackage{cuted}
\usepackage{capt-of}
\usepackage[hidelinks,colorlinks=false,pdfborder={0 0 0},breaklinks=true]{hyperref}
\usepackage{orcidlink}
\RequirePackage{iftex}
\ifPDFTeX\else
  \newcount\pdfcompresslevel
  \newcount\pdfoptionpdfminorversion
  \newcount\pdfgentounicode
  \providecommand{\pdfcatalog}[1]{}
  \providecommand{\pdfglyphtounicode}[2]{}
\fi

\usepackage{xcolor}
\definecolor{cmlabPrefix}{HTML}{0D207F}  
\definecolor{cmlabNumber}{HTML}{0D207F}  
\definecolor{cmlabCite}{HTML}{0D207F}    
\definecolor{cmlabURL}{HTML}{0D207F}     

\usepackage{hyperref}
\hypersetup{
  colorlinks=true,
  linkcolor=cmlabNumber,    
  citecolor=cmlabCite,      
  urlcolor=cmlabURL,        
  filecolor=cmlabURL,
  pdfborder={0 0 0},
}

\usepackage[capitalize,noabbrev]{cleveref}

\newcommand{\cmlabRef}[4]{%
  {\color{cmlabPrefix}#1}~#3{\color{cmlabNumber}#2}#4%
}
\newcommand{\cmlabRefParen}[4]{
  {\color{cmlabPrefix}#1}~#3{\color{cmlabNumber}(#2)}#4%
}

\crefformat{figure}     {\cmlabRef{Fig.}{#1}{#2}{#3}}
\Crefformat{figure}     {\cmlabRef{Figure}{#1}{#2}{#3}}
\crefrangeformat{figure}%
  {{\color{cmlabPrefix}Figs.}~#3{\color{cmlabNumber}#1}#4--#5{\color{cmlabNumber}#2}#6}
\crefmultiformat{figure}%
  {\cmlabRef{Figs.}{#1}{#2}{#3}}%
  {, #2{\color{cmlabNumber}#1}#3}%
  {, #2{\color{cmlabNumber}#1}#3}%
  {, and #2{\color{cmlabNumber}#1}#3}

\crefformat{table}      {\cmlabRef{Tab.}{#1}{#2}{#3}}
\Crefformat{table}      {\cmlabRef{Table}{#1}{#2}{#3}}
\crefrangeformat{table}%
  {{\color{cmlabPrefix}Tabs.}~#3{\color{cmlabNumber}#1}#4--#5{\color{cmlabNumber}#2}#6}
\crefmultiformat{table}%
  {\cmlabRef{Tabs.}{#1}{#2}{#3}}%
  {, #2{\color{cmlabNumber}#1}#3}%
  {, #2{\color{cmlabNumber}#1}#3}%
  {, and #2{\color{cmlabNumber}#1}#3}

\crefformat{section}    {\cmlabRef{Sec.}{#1}{#2}{#3}}
\Crefformat{section}    {\cmlabRef{Section}{#1}{#2}{#3}}
\crefformat{subsection} {\cmlabRef{Sec.}{#1}{#2}{#3}}
\Crefformat{subsection} {\cmlabRef{Section}{#1}{#2}{#3}}
\crefformat{subsubsection}{\cmlabRef{Sec.}{#1}{#2}{#3}}
\Crefformat{subsubsection}{\cmlabRef{Section}{#1}{#2}{#3}}
\crefrangeformat{section}%
  {{\color{cmlabPrefix}Secs.}~#3{\color{cmlabNumber}#1}#4--#5{\color{cmlabNumber}#2}#6}
\crefmultiformat{section}%
  {\cmlabRef{Secs.}{#1}{#2}{#3}}%
  {, #2{\color{cmlabNumber}#1}#3}%
  {, #2{\color{cmlabNumber}#1}#3}%
  {, and #2{\color{cmlabNumber}#1}#3}

\crefformat{equation}   {\cmlabRefParen{Eq.}{#1}{#2}{#3}}
\Crefformat{equation}   {\cmlabRefParen{Equation}{#1}{#2}{#3}}
\crefrangeformat{equation}%
  {{\color{cmlabPrefix}Eqs.}~#3{\color{cmlabNumber}(#1)}#4--#5{\color{cmlabNumber}(#2)}#6}
\crefmultiformat{equation}%
  {\cmlabRefParen{Eqs.}{#1}{#2}{#3}}%
  {, #2{\color{cmlabNumber}(#1)}#3}%
  {, #2{\color{cmlabNumber}(#1)}#3}%
  {, and #2{\color{cmlabNumber}(#1)}#3}

\crefname{algorithm}{Algo.}{Algos.}
\Crefname{algorithm}{Algorithm}{Algorithms}
\crefformat{algorithm}  {\cmlabRef{Algo.}{#1}{#2}{#3}}
\Crefformat{algorithm}  {\cmlabRef{Algorithm}{#1}{#2}{#3}}
\crefrangeformat{algorithm}%
  {{\color{cmlabPrefix}Algos.}~#3{\color{cmlabNumber}#1}#4--#5{\color{cmlabNumber}#2}#6}
\crefmultiformat{algorithm}%
  {\cmlabRef{Algos.}{#1}{#2}{#3}}%
  {, #2{\color{cmlabNumber}#1}#3}%
  {, #2{\color{cmlabNumber}#1}#3}%
  {, and #2{\color{cmlabNumber}#1}#3}

\makeatletter
\ifx\input@path\@undefined
  \def\input@path{{arXiv/}}
\else
  \edef\input@path{{arXiv/}\input@path}
\fi
\makeatother
\usepackage[
  enable,
  labname={CMLab},
  leftlogo={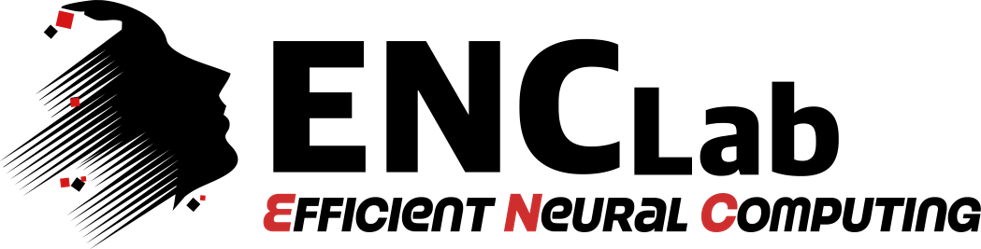},
  logo={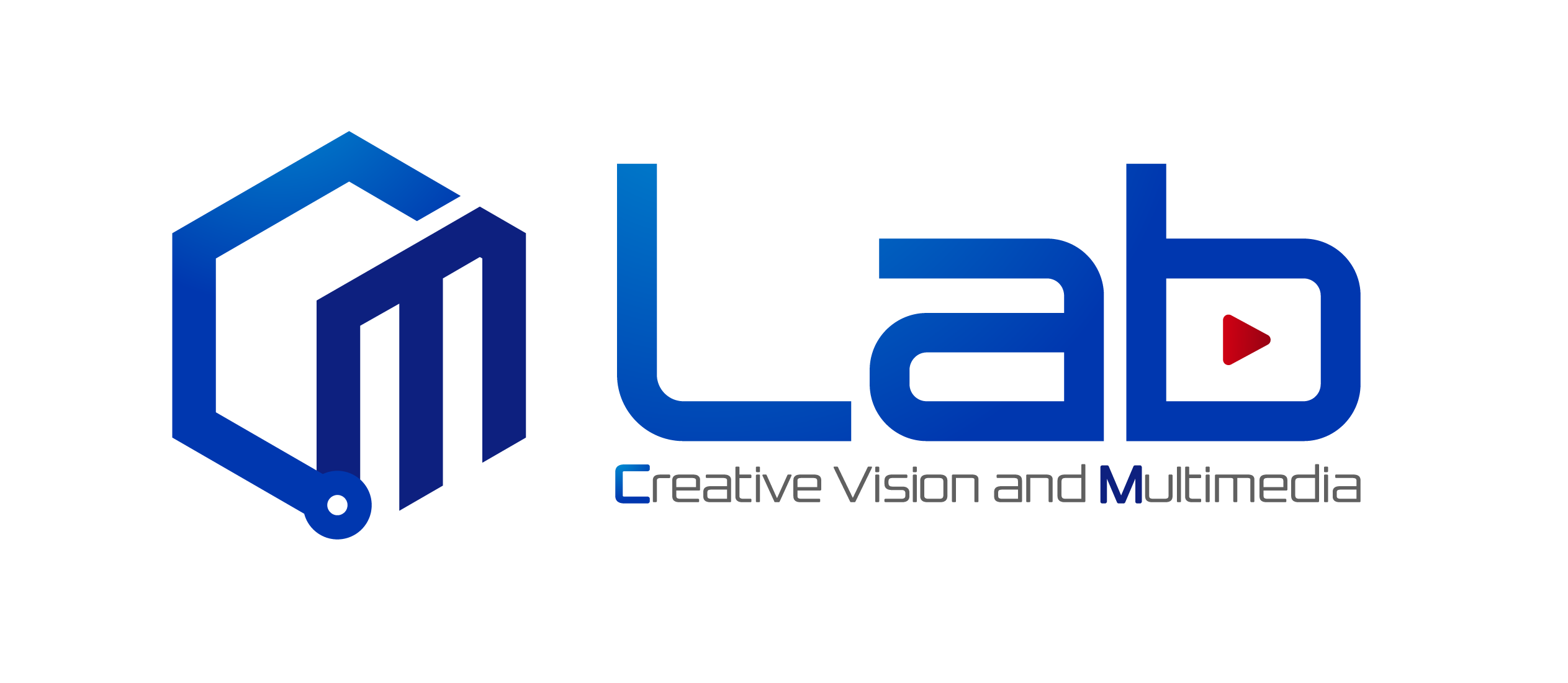},
  titleicon={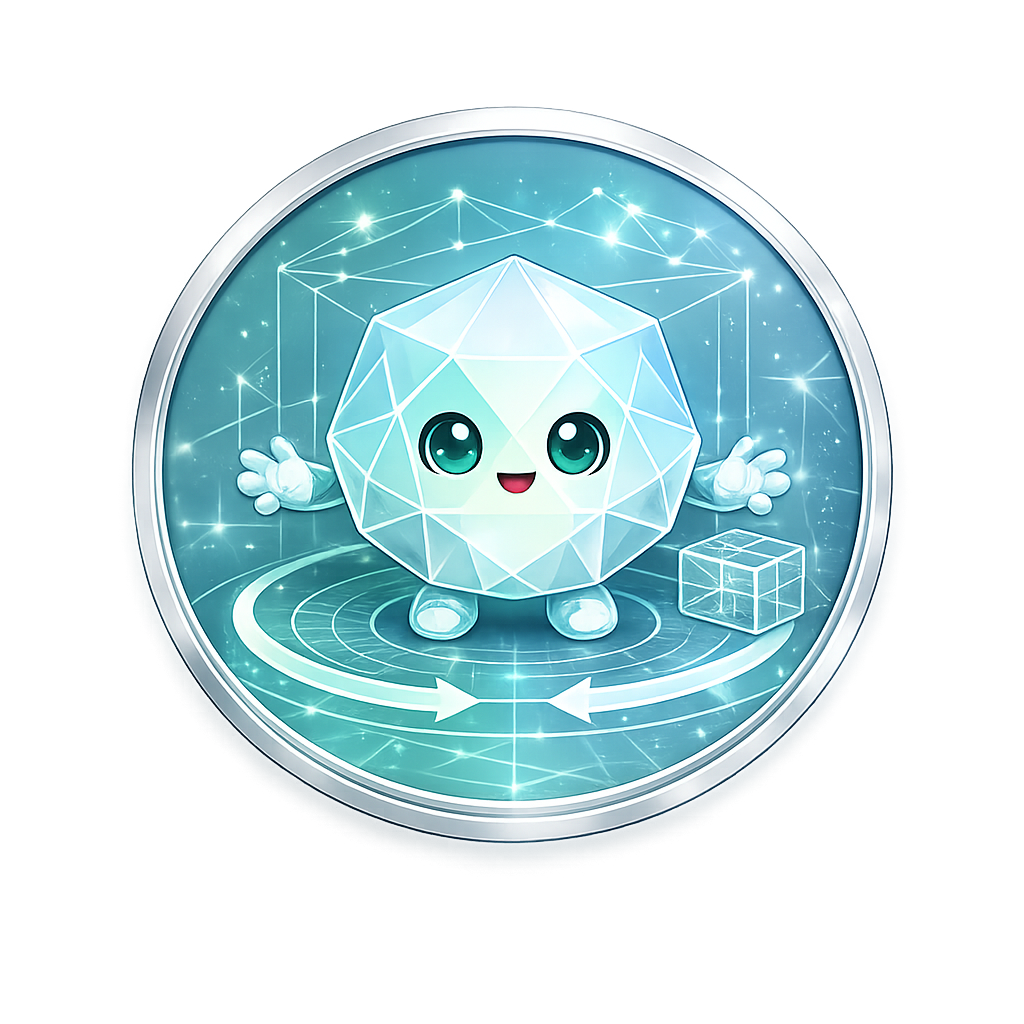},
  logowidth={22mm}
]{cmlab_1}

\newsavebox{\corrTableBox}
\newsavebox{\gdoTableBox}
\newsavebox{\fanTableBox}

\definecolor{first}{HTML}{C8E6C9}
\definecolor{second}{HTML}{FFF9C4}
\definecolor{third}{HTML}{BBDEFB}
\definecolor{checkgreen}{RGB}{0, 128, 128}
\definecolor{crossred}{RGB}{200, 30, 30}
\definecolor{triangleyellow}{RGB}{240, 170, 30}
\definecolor{skyblue}{RGB}{30, 144, 255}
\newcommand{\cmark}{\textcolor{checkgreen}{\ding{51}}}
\newcommand{\xmark}{\textcolor{crossred}{\ding{55}}}
\newcommand{\tmark}{\textcolor{triangleyellow}{$\blacktriangle$}}

\begin{document}

\title{Self-Geometry: GT-Free and Plug-and-Play Test-Time Adaptation for Geometrically Consistent 3D Vision Foundation Models}

\author{Seokhyun~Youn~\orcidlink{0009-0004-5265-2198},
        Dahyeon~Kye~\orcidlink{0009-0002-9233-195X},
        Sung-Ho~Bae~\orcidlink{0000-0002-3389-1159},~\IEEEmembership{Member,~IEEE,}
        and~Jihyong~Oh~\orcidlink{0000-0002-1627-0529},~\IEEEmembership{Member,~IEEE}
\thanks{Seokhyun Youn is with the CMLab, Department of Virtual Convergence, GSAIM, Chung-Ang University, Seoul, South Korea (e-mail: hisn16@cau.ac.kr).}%
\thanks{Dahyeon Kye is with the CMLab, Department of Imaging Science and Arts, GSAIM, Chung-Ang University, Seoul, South Korea (e-mail: rpekgus@cau.ac.kr).}%
\thanks{Sung-Ho Bae is with the School of Computing, Kyung Hee University, Yongin-si, South Korea (e-mail: shbae@khu.ac.kr).}%
\thanks{Jihyong Oh is with the CMLab, Department of Imaging Science and Arts, GSAIM, Chung-Ang University, Seoul, South Korea (e-mail: jihyongoh@cau.ac.kr).}%
\thanks{(\textit{Corresponding authors: Sung-Ho Bae; Jihyong Oh.})}%
\thanks{Manuscript received [Month Day, Year]; revised [Month Day, Year].}}

\pagestyle{plain}


\cmlabAuthors{Seokhyun Youn$^{1}$, Dahyeon Kye$^{1}$, Sung-Ho Bae$^{2,\dagger}$, Jihyong Oh$^{1,\dagger}$}
\cmlabAffiliations{$^{1}$CMLab, Chung-Ang University \quad $^{2}$Kyung Hee University \\ $^{\dagger}$Corresponding authors}
\cmlabAuthorEmail{\{hisn16, rpekgus, jihyongoh\}@cau.ac.kr \quad shbae@khu.ac.kr}
\cmlabProjectPage{https://cmlab-korea.github.io/Self-Geometry/}

\maketitle

\begin{strip}
\centering
\includegraphics[width=0.95\linewidth]{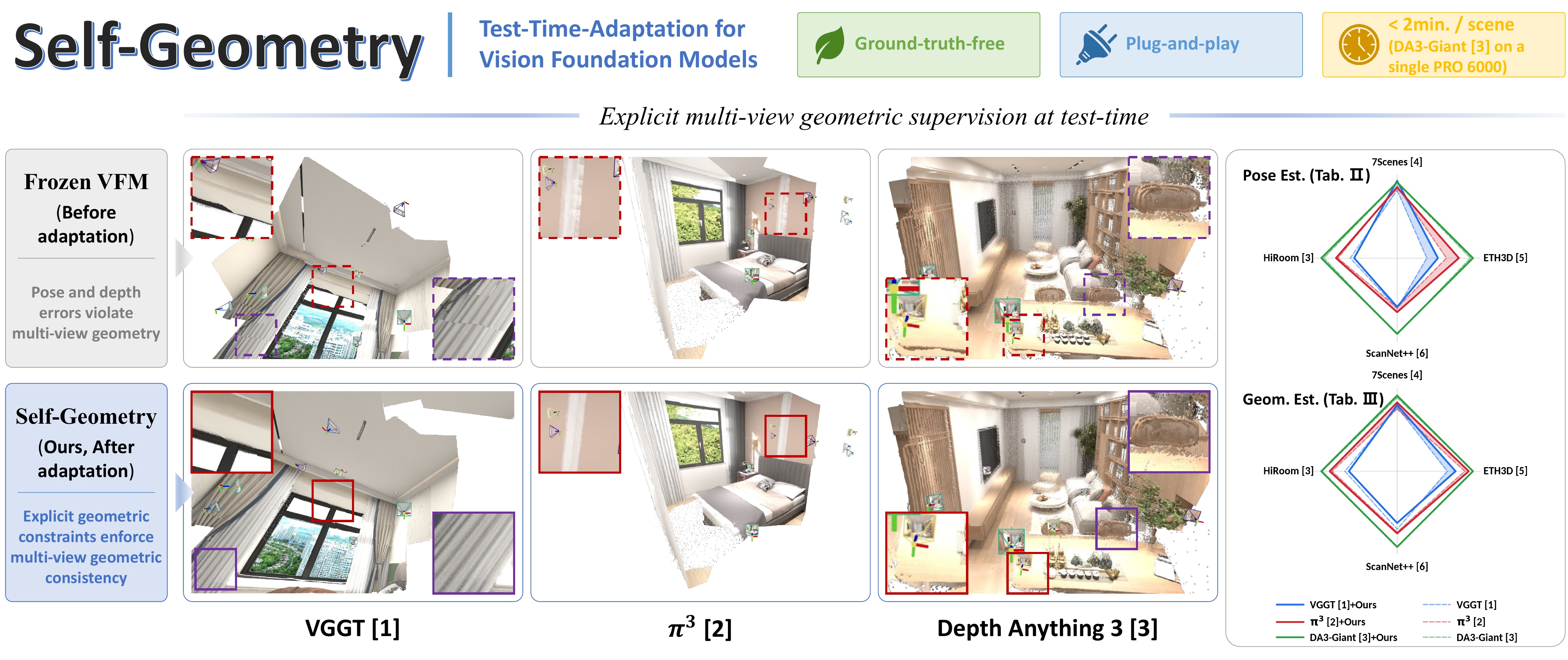}
\captionof{figure}{Overview and representative results of our proposed \textit{Self-Geometry}, a \textit{Plug-and-play} \textit{TTA} pipeline that imposes explicit multi-view geometric constraints on pretrained \textit{VFMs}. Our proposed \textit{Self-Geometry} yields consistent improvements in both Pose Estimation and Geometry Estimation across diverse pretrained \textit{VFMs} such as VGGT~\cite{wang2025vggt}, $\pi^3$~\cite{yang2025pi3}, and DA3-Giant/Large/Base/Small~\cite{lin2025da3}, and benchmark datasets including 7Scenes~\cite{shotton2013scenes}, ETH3D~\cite{schops2017eth3d}, ScanNet++~\cite{yeshwanth2023scannetpp}, and HiRoom~\cite{lin2025da3}.}
\label{fig:teaser}
\end{strip}

\begin{abstract}
Recent Vision Foundation Models (VFMs) predict depth, camera pose, and pointmap in a single forward pass without per-scene optimization, achieving strong generalization. However, enforcing explicit multi-view geometric consistency, e.g., through bundle adjustment, is computationally costly and is thus not imposed during VFM pretraining, so such inconsistency can arise. To address this, implicit self-consistency derived from model outputs (e.g., pointmaps, features), though enforced at test-time in prior work, delivers inherently limited performance gain, especially on scenes where the pretrained VFM is highly inaccurate. In contrast to this implicit signal, we propose \textit{Self-Geometry}, a plug-and-play test-time adaptation pipeline that directly imposes explicit multi-view geometric constraints using 2D pixel correspondences as pseudo ground-truth. Our proposed \textit{Self-Geometry} consists of Geometric Disentanglement Optimization, which combines Multi-View Consistency and Epipolar Consistency losses with Gradient Disentanglement to prevent gradient conflict; Frame Angular-Neighbor, a view sampler based on SO(3) geodesic distances for lightly imposing these constraints; and Lightweight TTA, which adapts VFMs via LoRA. Our method achieves consistent improvements in both pose and geometry estimation across six VFMs (VGGT, $\pi^3$, DA3-Giant/Large/Base/Small) and four benchmarks (7Scenes, ETH3D, ScanNet++, HiRoom).
\end{abstract}


\section{Introduction}
\label{sec:introduction}

\begingroup\color{black}

Multi-view 3D reconstruction aims to recover camera poses and dense scene geometry from multi-view images~\cite{hartley2003multiple}. Conventional methods~\cite{schoenberger2016colmap,triggs2000bundle} obtain accurate geometry through per-scene optimization but require substantial computation for every new scene. Recent \textit{Vision Foundation Models (VFMs)}, such as VGGT~\cite{wang2025vggt}, $\pi^3$~\cite{yang2025pi3}, and Depth Anything 3 (DA3)~\cite{lin2025da3}, instead predict depths, camera poses, and pointmaps in a single forward pass and achieve strong zero-shot performance across diverse visual geometry benchmarks, including 7Scenes~\cite{shotton2013scenes}, ScanNet++~\cite{yeshwanth2023scannetpp}, ETH3D~\cite{schops2017eth3d}, and HiRoom~\cite{lin2025da3}.

Despite this strong performance, enforcing explicit multi-view geometric consistency, e.g., through bundle adjustment~\cite{triggs2000bundle}, is prohibitively costly and thus omitted during \textit{VFM} pretraining~\cite{wang2025vggt}; multi-view geometric inconsistency can then arise in VFM predictions~\cite{bratulic2025geometric,selfi2025}. To address this inconsistency, the implicit self-consistency derived from model outputs (e.g., pointmaps, features), enforced at test-time in prior work~\cite{dai2026freegeometry}, improves the model only indirectly and lacks explicit geometric guarantee. As shown in Fig.~\ref{fig:failure}, Free-Geometry~\cite{dai2026freegeometry}, a representative implicit self-consistency method, barely improves camera pose or depth, and yields only marginal pointmap improvement, especially where the pretrained VFM is highly inaccurate.

\begin{figure*}[!t]
\centering
\includegraphics[width=\textwidth]{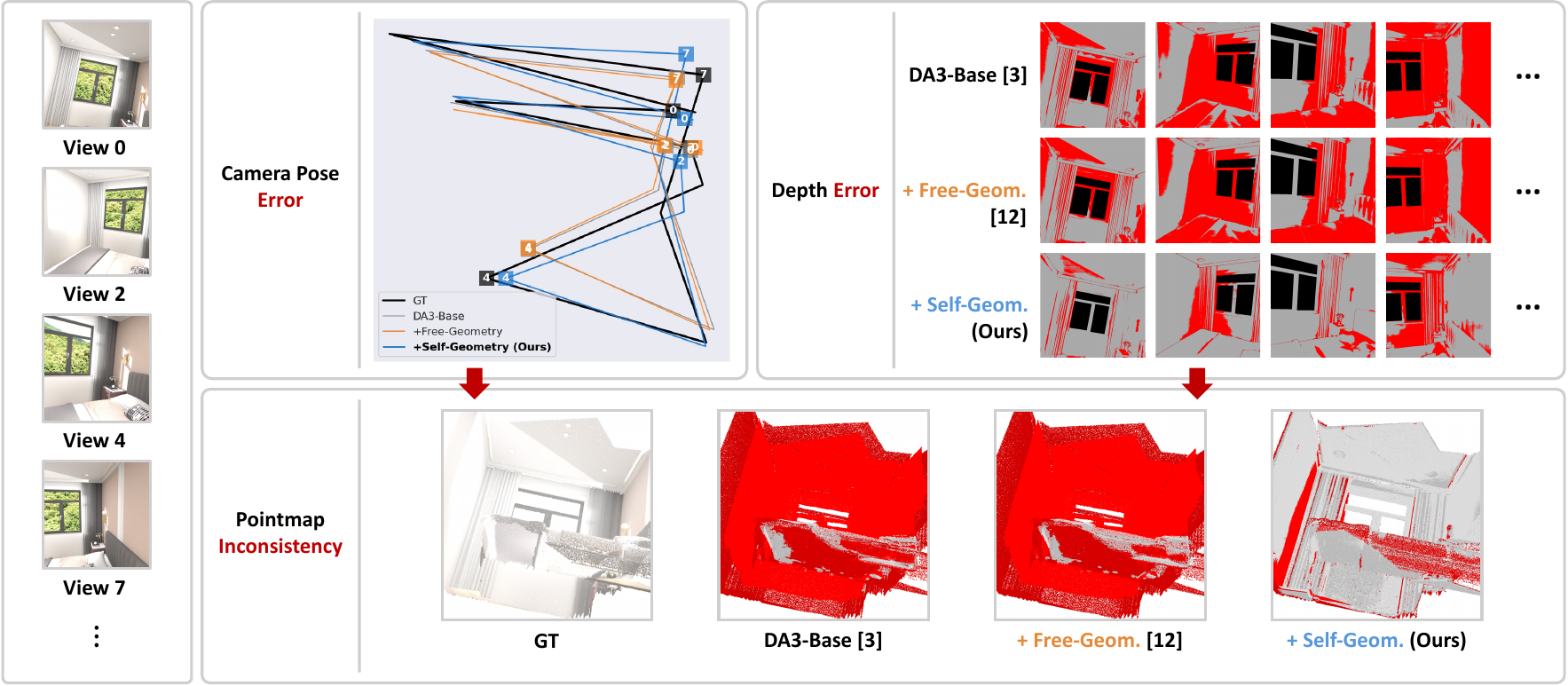}
\caption{Comparison of pretrained DA3-Base~\cite{lin2025da3} with two test-time adaptation methods on a HiRoom~\cite{lin2025da3} (\textit{828786}) scene, decomposed along camera pose error, depth error, and the resulting pointmap. +Free-Geometry~\cite{dai2026freegeometry}, a representative implicit self-consistency method, fails to substantially improve both the camera pose and depth, and consequently offers only limited improvement in the resulting pointmap. In contrast, \textcolor{blue}{+Self-Geometry} (Ours), which directly imposes explicit multi-view geometric constraints, yields consistent improvements across all three. In the depth error maps and pointmaps, \textcolor{red}{Red} regions denote prediction errors exceeding the benchmark threshold and \textcolor{gray}{gray} regions denote correctly reconstructed pixels.}
\label{fig:failure}
\end{figure*}

Since VFMs do not enforce this consistency during pretraining, we adopt \textit{Test-Time Adaptation (TTA)}~\cite{sun2020ttt,wang2021tent,wang2022cotta} to lightly impose explicit multi-view geometric constraints on the pretrained \textit{VFM} at test-time. Our key insight is that \textit{2D pixel correspondences}, extracted at test-time by an external feature matcher (\textit{e.g.}, LightGlue~\cite{lindenberger2023lightglue}), can serve as \textit{pseudo GT} for explicit multi-view geometric supervision~\cite{godard2019monodepth2,zhou2017sfmlearner}, eliminating the need for the GT annotations required by fine-tuning-based methods. These \textit{2D pixel correspondences} define the 2D pixel-level relations that camera poses and depths must satisfy~\cite{hartley2003multiple,longuethiggins1981}. Building on this insight, we propose \textit{Self-Geometry}, a \textit{Plug-and-play} \textit{TTA} pipeline that imposes explicit multi-view geometric constraints on pretrained \textit{VFMs}.

Under this pipeline, the proposed \textit{Multi-View Consistency Loss} (\textit{MVC Loss}; Eq.~\eqref{eq:mvc}) enforces such 2D pixel-level relations through a \textit{point-to-point} constraint that \textit{jointly} supervises the predicted camera poses and depths by penalizing reprojection errors using \textit{pseudo-correspondences} extracted at test-time. However, since the reprojection error depends on both the camera poses and the depths~\cite{triggs2000bundle}, the \textit{MVC Loss} inherits the well-known \textit{pose-depth ambiguity} of reprojection-based supervision. Specifically, distinct (pose, depth) pairs can yield the same \textit{MVC Loss} value, and thus the \textit{point-to-point} \textit{MVC Loss} alone cannot uniquely identify accurate pose or depth. To resolve this ambiguity, we introduce the \textit{Epipolar Consistency Loss} (\textit{EC Loss}; Eq.~\eqref{eq:ec}), a \textit{depth-independent} \textit{point-to-line} constraint that supervises the camera poses alone~\cite{hartley2003multiple}. Since both losses supervise the same camera poses, we further propose our \textit{Gradient Disentanglement} (\textit{GD}; Sec.~\ref{para:grad_disentangle}) at every \textit{TTA} iteration to prevent the \textit{gradient conflict} (Fig.~\ref{fig:ablation_gdo}(b)) that can arise between them. The proposed \textit{GD} projects the proposed \textit{MVC Loss} gradient onto the orthogonal complement of the proposed \textit{EC Loss} gradient~\cite{yu2020pcgrad}. As a result, the two losses operate complementarily: the \textit{point-to-line} \textit{EC Loss} refines camera poses independently of depth, and the \textit{point-to-point} \textit{MVC Loss} refines both at a fine-grained level unreachable by the proposed \textit{EC Loss}.

To lightly apply these explicit multi-view geometric constraints at test-time, we introduce two lightweight components. First, we introduce \textit{Frame Angular-Neighbor} (\textit{FAN}; Sec.~\ref{subsec:fan}), an SO(3)-guided~\cite{sola2018lie} view sampler that selects \textit{angularly diverse} source views using \textit{scene-scale-invariant} SO(3) geodesic distances~\cite{hartley2003multiple}, mitigating the quadratic cost of global attention in pretrained \textit{VFMs}. Second, we adopt \textit{Lightweight Test-Time Adaptation} (\textit{Lightweight TTA}; Sec.~\ref{subsec:lightweight_tta}) that inserts \textit{Low-Rank Adaptation (LoRA)}~\cite{hu2022lora} into the QKV weights of pretrained \textit{VFMs}' attention blocks (see Sec.~\ref{sec:supp_training} for per-VFM details). Together, these components enable our proposed \textit{Self-Geometry}, a \textit{Plug-and-play} \textit{TTA} pipeline imposing explicit multi-view geometric constraints on pretrained \textit{VFMs}, to complete per-scene adaptation within two minutes (up to 40 input views, DA3-Giant~\cite{lin2025da3} on ETH3D~\cite{schops2017eth3d}; Tab.~\ref{tab:complexity}) on a single NVIDIA RTX PRO 6000~\cite{nvidia_rtxpro6000}. Our contributions are summarized as follows:

\begin{itemize}
\item We propose \textit{Self-Geometry}, a \textit{GT-free} and \textit{Plug-and-play} \textit{TTA} pipeline that leverages \textit{2D pixel correspondences} as \textit{pseudo GT} to impose explicit multi-view geometric supervision on pretrained \textit{VFMs}.

\item Within our proposed \textit{Self-Geometry}, we design three complementary components: \textit{Geometric Disentanglement Optimization} (\textit{GDO}), which imposes explicit multi-view geometric supervision through \textit{GT-free} correspondence-guided losses; \textit{FAN}, an SO(3)-guided view sampler using \textit{scene-scale-invariant} SO(3) geodesic distances; and a \textit{LoRA}~\cite{hu2022lora}-based \textit{Lightweight TTA} strategy.

\item Across six pretrained \textit{VFMs} (VGGT~\cite{wang2025vggt}, $\pi^3$~\cite{yang2025pi3}, DA3-G/L/B/S~\cite{lin2025da3}) and four datasets (7Scenes~\cite{shotton2013scenes}, ETH3D~\cite{schops2017eth3d}, ScanNet++~\cite{yeshwanth2023scannetpp}, HiRoom~\cite{lin2025da3}), our proposed \textit{Self-Geometry} yields consistent improvements in both Pose Estimation and Geometry Estimation, completing scene-wise adaptation within two minutes per-scene on a single NVIDIA RTX PRO 6000~\cite{nvidia_rtxpro6000}.
\end{itemize}

\endgroup

\section{Related Work}
\label{sec:related_work}

\subsection{Frozen Vision Foundation Models}
\label{subsec:related_frozen}
Feed-forward 3D reconstruction has emerged as an alternative to per-scene SfM~\cite{schoenberger2016colmap}. Early such models operated on image pairs~\cite{wang2024dust3r,leroy2024mast3r} or streaming inputs~\cite{cut3r}. Recent \textit{VFMs} extend to arbitrary multi-view inputs: VGGT~\cite{wang2025vggt} uses alternating attention, $\pi^3$~\cite{yang2025pi3} uses permutation-equivariance, and DA3~\cite{lin2025da3} uses a unified depth-ray representation. Under a \textit{train-then-freeze} paradigm, these \textit{VFMs} are trained to regress camera pose and depth individually against \textit{GT annotations} without enforcing any explicit multi-view geometric consistency, so such inconsistency can arise in the resulting pointmap. Our proposed \textit{Self-Geometry} mitigates this inconsistency via \textit{GT-free} per-scene \textit{TTA}~\cite{sun2020ttt,wang2021tent,wang2022cotta}.

\subsection{Fine-tuning-based Adaptation}
\label{subsec:related_finetune}
Fine-tuning-based approaches modify pretrained \textit{VFMs} via additional training on external data or auxiliary priors~\cite{align3r,pow3r,mono3r,selfi2025}. Most relevant, Fin3R~\cite{fin3r} fine-tunes only the encoder of DUSt3R/MASt3R/CUT3R/VGGT~\cite{wang2024dust3r,leroy2024mast3r,cut3r,wang2025vggt} via \textit{LoRA}~\cite{hu2022lora} distillation from a monocular teacher. All fine-tune on a fixed dataset and cannot guarantee specialization to arbitrary target scenes. Our proposed \textit{Self-Geometry} performs per-scene \textit{TTA} using only pseudo-correspondences from each target scene, requiring no additional training dataset.

\subsection{Test-Time Adaptation}
\label{subsec:related_tta}
\textit{TTA}~\cite{yuan2025test3r,selfevo,ttt3r,online3r,lu2024lora3d,tco} specializes pretrained \textit{VFMs} using only the test scene. These methods rely on either auxiliary priors or implicit self-consistency supervision. TCO~\cite{tco} requires external priors such as camera poses, intrinsics, and depth. Free-Geometry~\cite{dai2026freegeometry} distills a full-view teacher into a masked-view student via \textit{LoRA}~\cite{hu2022lora} to enforce cross-view feature consistency. In contrast, our proposed \textit{Self-Geometry} lightly imposes explicit multi-view geometric constraints using only 2D pixel correspondences, without any auxiliary priors or teacher distillation.

\begin{table}[t]
\centering
\caption{Conceptual Comparisons of Multi-View 3D \textit{VFMs} Across Four Positioning Axes: \cmark denotes fully supported, \tmark denotes partially supported, and \xmark denotes unsupported.}
\label{tab:positioning}
\setlength{\tabcolsep}{2pt}
\renewcommand{\arraystretch}{0.95}
\resizebox{\columnwidth}{!}{%
\begin{tabular}{l|l|cccc}
\hline
\multicolumn{2}{c|}{Category / Method} & \shortstack{GT-\\free} & \shortstack{Teacher-\\free} & \shortstack{Plug-\&-\\play} & \shortstack{Explicit-\\Geometry} \\
\hline
(a) Frozen \textit{VFM}   & VGGT~\cite{wang2025vggt} / $\pi^3$~\cite{yang2025pi3} / DA3~\cite{lin2025da3}                       & \xmark & \tmark & ---    & \xmark \\
(b) Fine-tuning-based     & Pow3R~\cite{pow3r} / Fin3R~\cite{fin3r} / Selfi~\cite{selfi2025}  & \xmark & \xmark & \tmark & \tmark \\
\hline
\multirow{7}{*}{(c) \textit{TTA}-based}
                          & TCO~\cite{tco}                             & \xmark & \cmark & \cmark & \cmark \\
                          & Free-Geometry~\cite{dai2026freegeometry}   & \cmark & \xmark & \cmark & \xmark \\
                          & SelfEvo~\cite{selfevo}                     & \cmark & \xmark & \cmark & \xmark \\
                          & TTT3R~\cite{ttt3r}                         & \cmark & \cmark & \xmark & \xmark \\
                          & Online3R~\cite{online3r}                   & \cmark & \cmark & \xmark & \tmark \\
                          & Test3R~\cite{yuan2025test3r}               & \cmark & \cmark & \tmark & \tmark \\
                          & LoRA3D~\cite{lu2024lora3d}                 & \cmark & \cmark & \tmark & \cmark \\
\hline
\textbf{(d) Ours}         & \textit{\textbf{Self-Geometry}}            & \cmark & \cmark & \cmark & \cmark \\
\hline
\end{tabular}}
\end{table}

\subsection{Conceptual Positioning Across Four Axes}
\label{subsec:related_positioning}
As summarized in Tab.~\ref{tab:positioning}, existing approaches each exhibit at least one of the following four limitations. (i) \textit{GT-dependent}: Frozen \textit{VFMs} use \textit{GT annotations} during pretraining, and fine-tuning-based methods use them during additional training; collecting such GT annotations is costly. (ii) \textit{Teacher-dependent}: Fine-tuning-based methods, SelfEvo~\cite{selfevo}, and Free-Geometry~\cite{dai2026freegeometry} rely on \textit{teacher-model distillation}; student accuracy is upper-bounded by the teacher. (iii) \textit{Architecture-specific}: TTT3R~\cite{ttt3r}, Online3R~\cite{online3r}, Test3R~\cite{yuan2025test3r}, and LoRA3D~\cite{lu2024lora3d} are tied to \textit{specific architectures} such as CUT3R~\cite{cut3r}, MASt3R~\cite{leroy2024mast3r}, and DUSt3R~\cite{wang2024dust3r}, respectively; extending them to another VFM requires per-backbone redesign. (iv) \textit{Implicit-Geometry}: Free-Geometry, SelfEvo, and TTT3R rely on \textit{implicit self-consistency supervision} derived from model predictions such as pointmaps or features; this implicit signal delivers only limited performance gain. In contrast, our proposed \textit{Self-Geometry} enforces explicit geometric consistency, addressing all four axes simultaneously: \textit{GT-free}, \textit{Teacher-free}, \textit{Plug-and-play}, and \textit{Explicit-Geometry}.

\begin{figure*}[!t]
\centering
\includegraphics[width=\textwidth]{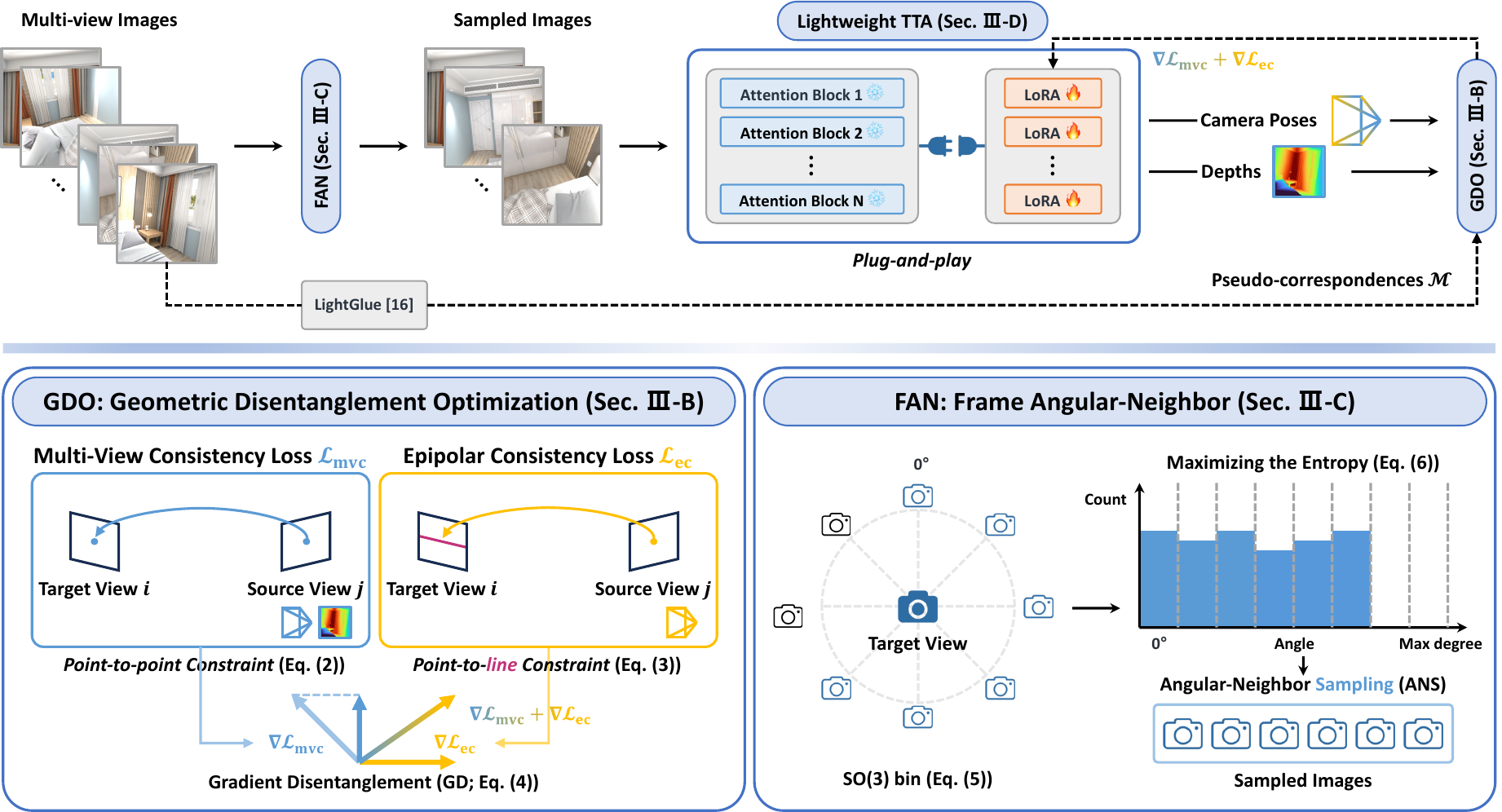}
\caption{Overview of Our Proposed \textit{Self-Geometry} Pipeline. \textit{Self-Geometry} adapts a frozen pretrained \textit{VFM} to a target scene through three complementary components. (i) \textit{GDO} (Sec.~\ref{subsec:gdo}) formulates two \textit{GT-free} correspondence-guided losses, the \textit{point-to-point} \textit{MVC Loss} (Eq.~\eqref{eq:mvc}) and the \textit{point-to-line} \textit{EC Loss} (Eq.~\eqref{eq:ec}), and applies \textit{GD} (Eq.~\eqref{eq:gdo_projection}) to prevent the \textit{gradient conflict} that can arise between them. (ii) \textit{FAN} (Sec.~\ref{subsec:fan}) samples angularly diverse views via \textit{scene-scale-invariant} SO(3) geodesic distances (Eq.~\eqref{eq:so3_angle}). (iii) \textit{Lightweight TTA} (Sec.~\ref{subsec:lightweight_tta}) updates only \textit{LoRA} (Sec.~\ref{para:lora_adapter}) parameters inserted into the attention blocks of the pretrained \textit{VFM}. Together, these components operate as a \textit{GT-free} and \textit{Plug-and-play} \textit{TTA} that imposes explicit multi-view geometric constraints on pretrained \textit{VFMs}.}
\label{fig:pipeline}
\end{figure*}

\section{Methodology}
\label{sec:method}

\subsection{Problem Setup and Key Intuition}
\label{subsec:setup}

\textbf{Problem Setup.} As discussed in Sec.~\ref{sec:introduction}, pretrained \textit{VFMs} are trained without enforcing any explicit multi-view geometric consistency, so such inconsistency can arise in the resulting pointmap (Fig.~\ref{fig:failure}). To resolve this inconsistency, we present our proposed \textit{Self-Geometry}, a \textit{GT-free} and \textit{Plug-and-play} \textit{TTA} pipeline that imposes explicit multi-view geometric constraints on a pretrained \textit{VFM}.

\textbf{Key Intuition: \textit{2D Pixel Correspondences Provide Explicit Multi-view Geometric Supervision}.} Our key insight is that \textit{2D pixel correspondences themselves define explicit multi-view geometric supervision} on the predictions of a pretrained \textit{VFM}.

\textbf{Pipeline Overview.} As depicted in Fig.~\ref{fig:pipeline}, our proposed \textit{Self-Geometry} consists of three complementary components. (i) \textit{Geometric Disentanglement Optimization} (\textit{GDO}; Sec.~\ref{subsec:gdo}) formulates the two constraints as the \textit{point-to-point} \textit{MVC Loss} and the \textit{point-to-line} \textit{EC Loss}, and applies the proposed \textit{GD} to prevent the \textit{gradient conflict} that can arise between them on the camera poses. (ii) \textit{Frame Angular-Neighbor} (\textit{FAN}; Sec.~\ref{subsec:fan}) samples input views using \textit{scene-scale-invariant} SO(3) geodesic distances, thereby providing geometrically rich supervision. (iii) \textit{Lightweight Test-Time Adaptation} (\textit{Lightweight TTA}; Sec.~\ref{subsec:lightweight_tta}) updates only lightweight \textit{LoRA}-based adapter parameters, keeping the pretrained \textit{VFM} frozen. Together, these components operate as our \textit{Self-Geometry}, a \textit{GT-free} and \textit{Plug-and-play} \textit{TTA} pipeline imposing explicit multi-view geometric constraints on pretrained \textit{VFMs}.

\subsection{Geometric Disentanglement Optimization (GDO)}
\label{subsec:gdo}

The proposed \textit{GDO} imposes explicit multi-view geometric constraints on pretrained \textit{VFMs} through the following five steps: (i) \textit{Scene Initialization} (Eq.~\eqref{eq:pseudo_corr_set}) extracts pseudo-correspondences from the input views using an external feature matcher; (ii) \textit{Pseudo-Correspondence Filtering} removes mismatched pseudo-correspondences; (iii) \textit{Multi-View Consistency Loss} (Eq.~\eqref{eq:mvc}) imposes a \textit{point-to-point} constraint that supervises both camera pose and depth; (iv) \textit{Epipolar Consistency Loss} (Eq.~\eqref{eq:ec}) imposes a \textit{depth-independent} \textit{point-to-line} constraint that supervises camera pose alone; and (v) \textit{Gradient Disentanglement} (Eq.~\eqref{eq:gdo_projection}) projects the proposed \textit{MVC Loss} gradient onto the orthogonal complement of the proposed \textit{EC Loss} gradient~\cite{yu2020pcgrad} to prevent the \textit{gradient conflict} that can arise between the two losses on the camera poses, so that the two losses operate complementarily.

\textbf{Scene Initialization.}\label{para:scene_init} In this step, at scene initialization, we apply an external feature matcher (\textit{i.e.}, LightGlue~\cite{lindenberger2023lightglue}) once to every pair of input views to extract \textit{2D pixel correspondences}, and \textit{reuse} the resulting 2D pixel correspondence set throughout all subsequent \textit{TTA} iterations.

Given the set of input views $\mathcal{I} = \{\mathbf{I}_i\}_{i=1}^{N}$, the pretrained \textit{VFM} predicts camera poses $\mathcal{P} = \{\mathbf{P}_i\}_{i=1}^{N}$, depth maps $\mathcal{D} = \{\mathbf{D}_i\}_{i=1}^{N}$, and pointmaps $\mathcal{X} = \{\mathbf{X}_i\}_{i=1}^{N}$ in a single forward pass. Here, $N \in \mathbb{N}$ is the number of input views and $\mathbf{I}_i \in \mathbb{R}^{H \times W \times 3}$ is the RGB image at the $i$-th view. The camera pose of each view $i$, $\mathbf{P}_i = (\mathbf{K}_i, \mathbf{R}_i, \mathbf{t}_i)$, consists of an intrinsic matrix $\mathbf{K}_i \in \mathbb{R}^{3\times 3}$, a rotation matrix $\mathbf{R}_i \in SO(3)$, and a translation vector $\mathbf{t}_i \in \mathbb{R}^{3}$, and $\mathbf{D}_i \in \mathbb{R}^{H \times W}$ and $\mathbf{X}_i \in \mathbb{R}^{H \times W \times 3}$ are the pixel-wise depth map and pointmap of view $i$, respectively. Fixing one view as the target view $i$ and the others as source views (each indexed by $j$), the pseudo-correspondence set from the external matcher is defined as:
\begin{equation}
\label{eq:pseudo_corr_set}
\mathcal{M} = \bigl\{ (\mathbf{x}_i^{m}, \mathbf{x}_j^{m}) \bigr\}_{m=1}^{M}.
\end{equation}
Here, $\mathbf{x}_i^{m}, \mathbf{x}_j^{m} \in \mathbb{R}^{2}$ are the pixel coordinates of the $m$-th pseudo-correspondence in the target and source views, and $M = |\mathcal{M}|$ is the total number of pseudo-correspondences. Each pair provides two distinct multi-view geometric constraints that the predicted $\mathcal{P}$ and $\mathcal{D}$ must satisfy, formalized as the \textit{point-to-point} \textit{MVC Loss} (Eq.~\eqref{eq:mvc}) and the \textit{point-to-line} \textit{EC Loss} (Eq.~\eqref{eq:ec}).

\textbf{Pseudo-Correspondence Filtering.}\label{para:filtering} The pseudo-correspondences extracted by the external matcher may contain mismatches that inject unreliable supervision. To remove them, before every \textit{TTA} iteration, we sequentially apply the proposed \textit{EC Loss}-based filter (Eq.~\eqref{eq:ec}), followed by the proposed \textit{MVC Loss}-based filter (Eq.~\eqref{eq:mvc}), to $\mathcal{M}$. The \textit{EC Loss}-based filter coarsely removes mismatches with large camera pose errors, and the proposed \textit{MVC Loss}-based filter eliminates the residual mismatches. We empirically validate the effectiveness of this filtering strategy in Sec.~\ref{para:ablation_filtering}.

\textbf{Multi-View Consistency Loss, $\mathcal{L}_{\mathrm{mvc}}$.}\label{para:mvc} The proposed \textit{MVC Loss} formalizes the \textit{point-to-point} reprojection constraint provided by the pseudo-correspondences. At each \textit{TTA} iteration, this loss constrains the camera poses and depths predicted by the pretrained \textit{VFM} to consistently reproject each 2D pixel in source view $j$ to its paired 2D pixel in target view $i$, as:
\begin{equation}
\label{eq:mvc}
\mathcal{L}_{\mathrm{mvc}}^{i}(\mathcal{P}, \mathcal{D}; \mathcal{M}) = \frac{1}{|\mathcal{M}|} \sum_{(\mathbf{x}_i^{m}, \mathbf{x}_j^{m}) \in \mathcal{M}} \left\|\hat{\mathbf{x}}_i^{m} - \mathbf{x}_i^{m}\right\|_2.
\end{equation}
Here, $\hat{\mathbf{x}}_i^{m}$ is the pixel obtained by reprojecting the pseudo-correspondence pixel $\mathbf{x}_j^{m}$ in source view $j$ into target view $i$ using the camera poses $\mathcal{P}$ and depth $\mathbf{D}_j$, and $\mathbf{x}_i^{m}$ is the pseudo-correspondence pixel in target view $i$. Since $\mathcal{L}_{\mathrm{mvc}}$ depends on both $\mathcal{P}$ and $\mathbf{D}_j$, it \textit{jointly} supervises the camera poses and depths. However, $\mathcal{L}_{\mathrm{mvc}}$ inherits the well-known \textit{pose-depth ambiguity} of reprojection-based supervision, where distinct $(\mathcal{P}, \mathbf{D}_j)$ pairs can yield the same $\mathcal{L}_{\mathrm{mvc}}$ value, and thus the \textit{point-to-point} \textit{MVC Loss} alone cannot uniquely identify accurate camera pose or depth.

\textbf{Epipolar Consistency Loss, $\mathcal{L}_{\mathrm{ec}}$.}\label{para:ec} To resolve this \textit{pose-depth ambiguity}, we further introduce the \textit{depth-independent} \textit{EC Loss} that formalizes the \textit{point-to-line} epipolar constraint provided by the pseudo-correspondences. At each \textit{TTA} iteration, this loss constrains the camera poses predicted by the pretrained \textit{VFM} to consistently place each 2D pixel in source view $j$ on the epipolar line of its paired 2D pixel in target view $i$, as:
\begin{equation}
\label{eq:ec}
\resizebox{0.9\columnwidth}{!}{$\displaystyle
\mathcal{L}_{\mathrm{ec}}^{i}(\mathcal{P}; \mathcal{M}) = \frac{1}{|\mathcal{M}|} \sum_{(\mathbf{x}_i^{m}, \mathbf{x}_j^{m}) \in \mathcal{M}} \sqrt{d_{\mathrm{Sampson}}\!\left(\tilde{\mathbf{x}}_i^{m}, \tilde{\mathbf{x}}_j^{m}; \mathbf{F}_{i\leftarrow j}\right)}.
$}
\end{equation}
Here, $\mathbf{F}_{i\leftarrow j}$ is the fundamental matrix from source view $j$ to target view $i$ derived from the camera poses $\mathcal{P}$ predicted by the pretrained \textit{VFM}, $d_{\mathrm{Sampson}}(\cdot,\cdot;\mathbf{F})$ is the Sampson distance (see Sec.~\ref{sec:preliminary}), and $\tilde{\mathbf{x}}_i^{m}$ and $\tilde{\mathbf{x}}_j^{m}$ are the homogeneous coordinates of the pseudo-correspondence pixels $\mathbf{x}_i^{m}$ and $\mathbf{x}_j^{m}$. $\mathcal{L}_{\mathrm{ec}}$ depends only on $\mathcal{P}$, supervising the camera poses in a \textit{depth-independent} manner.

\textbf{Gradient Disentanglement.}\label{para:grad_disentangle} Since both losses supervise the same camera poses, we introduce \textit{Gradient Disentanglement} (\textit{GD}) at every \textit{TTA} iteration to prevent the \textit{gradient conflict} (Fig.~\ref{fig:ablation_gdo}(b)) that can arise between them, projecting $\nabla \mathcal{L}_{\mathrm{mvc}}$ onto the orthogonal complement of $\nabla \mathcal{L}_{\mathrm{ec}}$, as:
\begin{equation}
\label{eq:gdo_projection}
\nabla \mathcal{L}_{\mathrm{mvc}} \;\leftarrow\; \nabla \mathcal{L}_{\mathrm{mvc}} - \frac{\left\langle \nabla \mathcal{L}_{\mathrm{mvc}},\; \nabla \mathcal{L}_{\mathrm{ec}} \right\rangle}{\left\|\nabla \mathcal{L}_{\mathrm{ec}}\right\|_2^{2}}\, \nabla \mathcal{L}_{\mathrm{ec}}.
\end{equation}
Here, $\nabla \mathcal{L}_{\mathrm{ec}}$ and $\nabla \mathcal{L}_{\mathrm{mvc}}$ are the gradients of $\mathcal{L}_{\mathrm{ec}}$ and $\mathcal{L}_{\mathrm{mvc}}$ with respect to the learnable parameters, and $\langle \cdot, \cdot \rangle$ denotes the inner product between the flattened gradient vectors. This projection preserves $\nabla \mathcal{L}_{\mathrm{ec}}$ and removes only the $\nabla \mathcal{L}_{\mathrm{ec}}$-direction component from $\nabla \mathcal{L}_{\mathrm{mvc}}$, so that the two losses operate complementarily at the gradient level: the \textit{point-to-line} \textit{EC Loss} refines the camera poses independently of the depths, and the \textit{point-to-point} \textit{MVC Loss} refines the camera poses and depths at a fine-grained level that the proposed \textit{EC Loss} cannot reach.

\subsection{Frame Angular-Neighbor (FAN)}
\label{subsec:fan}

As discussed in Sec.~\ref{sec:introduction}, to lightly apply these explicit multi-view geometric constraints, we need an efficient view sampling strategy. To minimize information loss about the target scene induced by the sampled view set, it must cover the scene evenly with moderate inter-frame overlap, which depends on the relative displacement between the sampled cameras. Although SE(3)~\cite{sola2018lie} geodesic distance measures this displacement fully, its rotation and translation components have distinct units and its translation is \textit{scene-scale-variant}, requiring heuristic per-scene normalization that unnecessarily complicates the view sampling pipeline. We therefore introduce \textit{FAN}, which uses SO(3) geodesic distance alone; since relative rotation primarily determines viewing direction and thus scene coverage, this \textit{scene-scale-invariant} measure can achieve uniform scene coverage without the translation component. At each \textit{TTA} iteration, the proposed \textit{FAN} samples angularly diverse views using SO(3) geodesic distances (\textit{i.e.}, the relative rotation angles), and proceeds in two steps: (i) \textit{Geometry-Rich View Selection} (\textit{GRV}; Eq.~\eqref{eq:grv_selection}) selects the target view based on SO(3) geodesic diversity, and (ii) \textit{Angular-Neighbor Sampling} (\textit{ANS}) samples source views from each SO(3) bin defined relative to it.

\textbf{Geometry-Rich View Selection (GRV).}\label{para:grv} The proposed \textit{GRV} selects the target view in two steps and uses SO(3) bins constructed from the SO(3) geodesic distance between input views. The SO(3) geodesic distance between two input views $i$ and $j$ is defined as:
\begin{equation}
\label{eq:so3_angle}
\theta_{ij} = \cos^{-1}\!\left(\frac{\mathrm{tr}(\mathbf{R}_i \mathbf{R}_j^{\top}) - 1}{2}\right).
\end{equation}
Here, $\mathbf{R}_i, \mathbf{R}_j \in SO(3)$ are the rotation matrices of the respective views, $\mathrm{tr}(\cdot)$ is the trace operator, and $\theta_{ij} \in [0, \pi]$ is the SO(3) geodesic distance between the two rotation matrices. We partition the value range $[0, \pi]$ of $\theta_{ij}$ into $B$ disjoint intervals to construct the SO(3) bins, and each view is assigned to the bin whose interval contains its $\theta_{ij}$.

In the first step, each input view $v$ is treated as a candidate target view, and the remaining views are assigned as source views to the SO(3) bins; a bin with at least one assigned source view is defined as an \textit{active SO(3) bin}. Since the number of active SO(3) bins reflects the scene coverage of the source views with respect to the target view, we obtain the candidate target view index set $\mathcal{V} \subseteq \{1, \dots, N\}$ maximizing this count.

In the second step, from the candidates in $\mathcal{V}$, the view that maximizes the SO(3) bin entropy is selected as the target view:
\begin{equation}
\label{eq:grv_selection}
\begin{gathered}
v^{*} = \arg\max_{v \in \mathcal{V}} H(v), \quad H(v) = -\sum_{b=1}^{B} p_b(v) \log p_b(v), \\
p_b(v) = \frac{n_b(v)}{\sum_{k=1}^{B} n_k(v)}.
\end{gathered}
\end{equation}
Here, $n_b(v)$ is the number of source views assigned to bin $b$ for candidate target view $v$. In Eq.~\eqref{eq:grv_selection}, maximizing the entropy $H(v)$ corresponds to selecting a target view $v^{*}$ whose source views are uniformly distributed across all active SO(3) bins rather than concentrated in specific angular regions. Through this two-step procedure, the sampled view set at each \textit{TTA} iteration achieves uniform scene coverage.

\textbf{Angular-Neighbor Sampling (ANS).}\label{para:ans} The proposed \textit{ANS} samples source views from each SO(3) bin defined with respect to the target view $v^{*}$ selected once by the proposed \textit{GRV} at scene initialization, constructing a compact sampled view set at each \textit{TTA} iteration. This bin-wise sampling prevents source views from being concentrated in specific angular regions and ensures uniform scene coverage.

\subsection{Lightweight Test-Time Adaptation (Lightweight TTA)}
\label{subsec:lightweight_tta}

The proposed \textit{Lightweight TTA} is a parameter-efficient optimization component. It specializes the pretrained \textit{VFM} to the target scene without additional training data, preserving the geometric prior acquired during pretraining. It employs a \textit{LoRA}~\cite{hu2022lora}-based adapter and updates only its parameters while keeping the pretrained \textit{VFM} frozen. It works with the proposed \textit{FAN} (Sec.~\ref{subsec:fan}) to complete per-scene adaptation on a single GPU within two minutes.

\textbf{LoRA-Based Adapter.}\label{para:lora_adapter} Our proposed \textit{LoRA-based adapter} targets the attention blocks of the pretrained \textit{VFM}. Specifically, following Fin3R~\cite{fin3r}, which \textit{LoRA} fine-tunes pretrained \textit{VFMs}, we insert \textit{LoRA}~\cite{hu2022lora} only into the QKV weights of these blocks, keeping all parameters of the pretrained \textit{VFM} frozen and updating only the \textit{LoRA} parameters.

Together, the proposed \textit{GDO}, \textit{FAN}, and \textit{Lightweight TTA} form our \textit{Self-Geometry}, a \textit{GT-free} and \textit{Plug-and-play} \textit{TTA} pipeline imposing explicit multi-view geometric constraints on pretrained \textit{VFMs}, adapting each scene on a single NVIDIA RTX PRO 6000~\cite{nvidia_rtxpro6000} within two minutes. The total loss formulation and algorithm are provided in Sec.~\ref{sec:supp_total_loss} and Sec.~\ref{sec:supp_algorithm}, respectively.

\section{Experiments}
\label{sec:experiments}

\begin{table*}[!t]
\centering
\caption{Pose Estimation results: AUC@3, AUC@30 ($\uparrow$) on 7Scenes~\cite{shotton2013scenes}, ETH3D~\cite{schops2017eth3d}, ScanNet++~\cite{yeshwanth2023scannetpp}, HiRoom~\cite{lin2025da3}, and Mean. Row notation: DA3-G/L/B/S denote DA3-Giant/Large/Base/Small~\cite{lin2025da3}; each pretrained model (e.g., VGGT, $\pi^3$, DA3-G) appears in four consecutive rows --- the baseline and three adapted variants with suffix ``-TCO'' (\textit{TCO}~\cite{tco}), ``-Fr'' (\textit{Free-Geometry}~\cite{dai2026freegeometry}), and ``-Ours'' (our proposed \textit{Self-Geometry}). $\Delta$\% vs.\ baseline in parentheses. Per column, \textcolor{blue}{blue}/\textcolor{red}{red} marks the largest gain/degradation among the three adaptation methods.}
\label{tab:pose_main}
{%
\resizebox{\textwidth}{!}{
\begin{tabular}{l|ll|ll|ll|ll|ll}
\hline
\multirow{2}{*}{Method} & \multicolumn{2}{c|}{7Scenes} & \multicolumn{2}{c|}{ETH3D} & \multicolumn{2}{c|}{ScanNet++} & \multicolumn{2}{c|}{HiRoom} & \multicolumn{2}{c}{\textcolor{skyblue}{Mean}} \\
 & AUC@3 $\uparrow$ & AUC@30 $\uparrow$ & AUC@3 $\uparrow$ & AUC@30 $\uparrow$ & AUC@3 $\uparrow$ & AUC@30 $\uparrow$ & AUC@3 $\uparrow$ & AUC@30 $\uparrow$ & AUC@3 $\uparrow$ & AUC@30 $\uparrow$ \\
\hline
VGGT & 0.25\phantom{ (\textcolor{blue}{\textbf{+15.1\%}})} & 0.86\phantom{ (\textcolor{blue}{\textbf{+15.1\%}})} & 0.20\phantom{ (\textcolor{blue}{\textbf{+15.1\%}})} & 0.76\phantom{ (\textcolor{blue}{\textbf{+15.1\%}})} & 0.56\phantom{ (\textcolor{blue}{\textbf{+15.1\%}})} & 0.94\phantom{ (\textcolor{blue}{\textbf{+15.1\%}})} & 0.51\phantom{ (\textcolor{blue}{\textbf{+15.1\%}})} & 0.88\phantom{ (\textcolor{blue}{\textbf{+15.1\%}})} & \textbf{0.38}\phantom{ (\textcolor{blue}{\textbf{+15.1\%}})} & \textbf{0.86}\phantom{ (\textcolor{blue}{\textbf{+15.1\%}})} \\
VGGT\textbf{-TCO} & 0.26 (\textbf{+4.3\%}) & 0.86 (\textbf{+0.3\%}) & 0.17 (\textcolor{red}{\textbf{-12.2\%}}) & 0.72 (\textcolor{red}{\textbf{-5.4\%}}) & 0.43 (\textcolor{red}{\textbf{-21.9\%}}) & 0.92 (\textcolor{red}{\textbf{-1.7\%}}) & 0.42 (\textcolor{red}{\textbf{-18.4\%}}) & 0.86 (\textbf{-1.7\%}) & \textbf{0.32} (\textcolor{red}{\textbf{-15.2\%}}) & \textbf{0.84} (\textcolor{red}{\textbf{-2.0\%}}) \\
VGGT\textbf{-Fr} & 0.25 (\textcolor{red}{\textbf{-0.2\%}}) & 0.86 (\textcolor{red}{\textbf{-0.0\%}}) & 0.20 (\textbf{+3.9\%}) & 0.77 (\textbf{+1.1\%}) & 0.56 (\textcolor{blue}{\textbf{+0.1\%}}) & 0.94 (\textcolor{blue}{\textbf{+0.0\%}}) & 0.54 (\textcolor{blue}{\textbf{+4.8\%}}) & 0.89 (\textcolor{blue}{\textbf{+1.8\%}}) & \textbf{0.39} (\textbf{+2.1\%}) & \textbf{0.86} (\textbf{+0.7\%}) \\
VGGT\textbf{-Ours} & 0.28 (\textcolor{blue}{\textbf{+13.9\%}}) & 0.87 (\textcolor{blue}{\textbf{+1.1\%}}) & 0.27 (\textcolor{blue}{\textbf{+37.3\%}}) & 0.83 (\textcolor{blue}{\textbf{+9.2\%}}) & 0.54 (\textbf{-2.9\%}) & 0.94 (\textbf{-0.0\%}) & 0.47 (\textbf{-8.2\%}) & 0.85 (\textcolor{red}{\textbf{-3.4\%}}) & \textbf{0.39} (\textcolor{blue}{\textbf{+3.3\%}}) & \textbf{0.87} (\textcolor{blue}{\textbf{+1.4\%}}) \\
\hline
$\pi^3$ & 0.26\phantom{ (\textcolor{blue}{\textbf{+15.1\%}})} & 0.86\phantom{ (\textcolor{blue}{\textbf{+15.1\%}})} & 0.33\phantom{ (\textcolor{blue}{\textbf{+15.1\%}})} & 0.86\phantom{ (\textcolor{blue}{\textbf{+15.1\%}})} & 0.55\phantom{ (\textcolor{blue}{\textbf{+15.1\%}})} & 0.94\phantom{ (\textcolor{blue}{\textbf{+15.1\%}})} & 0.65\phantom{ (\textcolor{blue}{\textbf{+15.1\%}})} & 0.94\phantom{ (\textcolor{blue}{\textbf{+15.1\%}})} & \textbf{0.45}\phantom{ (\textcolor{blue}{\textbf{+15.1\%}})} & \textbf{0.90}\phantom{ (\textcolor{blue}{\textbf{+15.1\%}})} \\
$\pi^3$\textbf{-TCO} & 0.02 (\textcolor{red}{\textbf{-90.5\%}}) & 0.42 (\textcolor{red}{\textbf{-51.3\%}}) & 0.09 (\textcolor{red}{\textbf{-72.0\%}}) & 0.60 (\textcolor{red}{\textbf{-30.2\%}}) & 0.02 (\textcolor{red}{\textbf{-96.0\%}}) & 0.63 (\textcolor{red}{\textbf{-32.5\%}}) & 0.01 (\textcolor{red}{\textbf{-97.8\%}}) & 0.51 (\textcolor{red}{\textbf{-46.1\%}}) & \textbf{0.04} (\textcolor{red}{\textbf{-91.5\%}}) & \textbf{0.54} (\textcolor{red}{\textbf{-40.0\%}}) \\
$\pi^3$\textbf{-Fr} & 0.26 (\textbf{-0.3\%}) & 0.86 (\textbf{+0.0\%}) & 0.32 (\textbf{-3.4\%}) & 0.85 (\textbf{-0.1\%}) & 0.45 (\textbf{-18.4\%}) & 0.93 (\textbf{-1.3\%}) & 0.63 (\textbf{-2.1\%}) & 0.95 (\textbf{+0.6\%}) & \textbf{0.41} (\textbf{-7.1\%}) & \textbf{0.90} (\textbf{-0.2\%}) \\
$\pi^3$\textbf{-Ours} & 0.26 (\textbf{-1.5\%}) & 0.86 (\textcolor{blue}{\textbf{+0.3\%}}) & 0.41 (\textcolor{blue}{\textbf{+25.1\%}}) & 0.90 (\textcolor{blue}{\textbf{+5.0\%}}) & 0.60 (\textcolor{blue}{\textbf{+8.8\%}}) & 0.95 (\textcolor{blue}{\textbf{+1.6\%}}) & 0.67 (\textcolor{blue}{\textbf{+3.4\%}}) & 0.95 (\textcolor{blue}{\textbf{+0.8\%}}) & \textbf{0.48} (\textcolor{blue}{\textbf{+8.3\%}}) & \textbf{0.92} (\textcolor{blue}{\textbf{+1.9\%}}) \\
\hline
DA3-G & 0.27\phantom{ (\textcolor{blue}{\textbf{+15.1\%}})} & 0.87\phantom{ (\textcolor{blue}{\textbf{+15.1\%}})} & 0.49\phantom{ (\textcolor{blue}{\textbf{+15.1\%}})} & 0.91\phantom{ (\textcolor{blue}{\textbf{+15.1\%}})} & 0.85\phantom{ (\textcolor{blue}{\textbf{+15.1\%}})} & 0.98\phantom{ (\textcolor{blue}{\textbf{+15.1\%}})} & 0.80\phantom{ (\textcolor{blue}{\textbf{+15.1\%}})} & 0.96\phantom{ (\textcolor{blue}{\textbf{+15.1\%}})} & \textbf{0.60}\phantom{ (\textcolor{blue}{\textbf{+15.1\%}})} & \textbf{0.93}\phantom{ (\textcolor{blue}{\textbf{+15.1\%}})} \\
DA3-G\textbf{-TCO} & 0.28 (\textcolor{blue}{\textbf{+1.6\%}}) & 0.87 (\textcolor{blue}{\textbf{+0.1\%}}) & 0.50 (\textbf{+2.5\%}) & 0.92 (\textbf{+1.0\%}) & 0.84 (\textcolor{red}{\textbf{-1.3\%}}) & 0.98 (\textcolor{red}{\textbf{-0.2\%}}) & 0.81 (\textbf{+1.6\%}) & 0.96 (\textbf{+0.1\%}) & \textbf{0.61} (\textbf{+0.8\%}) & \textbf{0.93} (\textbf{+0.2\%}) \\
DA3-G\textbf{-Fr} & 0.28 (\textbf{+0.8\%}) & 0.87 (\textbf{+0.1\%}) & 0.52 (\textcolor{blue}{\textbf{+7.9\%}}) & 0.92 (\textcolor{blue}{\textbf{+1.1\%}}) & 0.85 (\textcolor{blue}{\textbf{+0.2\%}}) & 0.98 (\textcolor{blue}{\textbf{+0.0\%}}) & 0.82 (\textbf{+1.9\%}) & 0.98 (\textcolor{blue}{\textbf{+1.7\%}}) & \textbf{0.62} (\textcolor{blue}{\textbf{+2.4\%}}) & \textbf{0.94} (\textcolor{blue}{\textbf{+0.7\%}}) \\
DA3-G\textbf{-Ours} & 0.27 (\textbf{+0.0\%}) & 0.87 (\textbf{+0.0\%}) & 0.50 (\textbf{+3.4\%}) & 0.91 (\textcolor{red}{\textbf{-0.2\%}}) & 0.84 (\textbf{-1.0\%}) & 0.98 (\textbf{-0.1\%}) & 0.83 (\textcolor{blue}{\textbf{+3.5\%}}) & 0.96 (\textbf{+0.4\%}) & \textbf{0.61} (\textbf{+1.5\%}) & \textbf{0.93} (\textbf{+0.1\%}) \\
\hline
DA3-L & 0.29\phantom{ (\textcolor{blue}{\textbf{+15.1\%}})} & 0.86\phantom{ (\textcolor{blue}{\textbf{+15.1\%}})} & 0.32\phantom{ (\textcolor{blue}{\textbf{+15.1\%}})} & 0.87\phantom{ (\textcolor{blue}{\textbf{+15.1\%}})} & 0.56\phantom{ (\textcolor{blue}{\textbf{+15.1\%}})} & 0.94\phantom{ (\textcolor{blue}{\textbf{+15.1\%}})} & 0.59\phantom{ (\textcolor{blue}{\textbf{+15.1\%}})} & 0.94\phantom{ (\textcolor{blue}{\textbf{+15.1\%}})} & \textbf{0.44}\phantom{ (\textcolor{blue}{\textbf{+15.1\%}})} & \textbf{0.90}\phantom{ (\textcolor{blue}{\textbf{+15.1\%}})} \\
DA3-L\textbf{-TCO} & 0.30 (\textcolor{blue}{\textbf{+2.6\%}}) & 0.86 (\textcolor{blue}{\textbf{+0.1\%}}) & 0.31 (\textcolor{red}{\textbf{-3.7\%}}) & 0.87 (\textcolor{red}{\textbf{-0.2\%}}) & 0.54 (\textcolor{red}{\textbf{-2.9\%}}) & 0.94 (\textcolor{red}{\textbf{-0.3\%}}) & 0.62 (\textbf{+5.0\%}) & 0.95 (\textcolor{blue}{\textbf{+0.6\%}}) & \textbf{0.44} (\textbf{+0.5\%}) & \textbf{0.90} (\textbf{+0.1\%}) \\
DA3-L\textbf{-Fr} & 0.29 (\textbf{+0.4\%}) & 0.86 (\textbf{-0.1\%}) & 0.36 (\textcolor{blue}{\textbf{+13.3\%}}) & 0.88 (\textcolor{blue}{\textbf{+1.6\%}}) & 0.56 (\textcolor{blue}{\textbf{+0.9\%}}) & 0.94 (\textcolor{blue}{\textbf{+0.1\%}}) & 0.58 (\textcolor{red}{\textbf{-1.0\%}}) & 0.94 (\textbf{+0.1\%}) & \textbf{0.45} (\textcolor{blue}{\textbf{+2.5\%}}) & \textbf{0.91} (\textcolor{blue}{\textbf{+0.4\%}}) \\
DA3-L\textbf{-Ours} & 0.29 (\textcolor{red}{\textbf{-0.3\%}}) & 0.86 (\textcolor{red}{\textbf{-0.1\%}}) & 0.32 (\textbf{-1.5\%}) & 0.87 (\textbf{-0.2\%}) & 0.56 (\textbf{+0.0\%}) & 0.94 (\textbf{+0.0\%}) & 0.62 (\textcolor{blue}{\textbf{+5.6\%}}) & 0.94 (\textbf{+0.4\%}) & \textbf{0.45} (\textbf{+1.6\%}) & \textbf{0.90} (\textbf{+0.0\%}) \\
\hline
DA3-B & 0.21\phantom{ (\textcolor{blue}{\textbf{+15.1\%}})} & 0.82\phantom{ (\textcolor{blue}{\textbf{+15.1\%}})} & 0.15\phantom{ (\textcolor{blue}{\textbf{+15.1\%}})} & 0.75\phantom{ (\textcolor{blue}{\textbf{+15.1\%}})} & 0.20\phantom{ (\textcolor{blue}{\textbf{+15.1\%}})} & 0.81\phantom{ (\textcolor{blue}{\textbf{+15.1\%}})} & 0.19\phantom{ (\textcolor{blue}{\textbf{+15.1\%}})} & 0.83\phantom{ (\textcolor{blue}{\textbf{+15.1\%}})} & \textbf{0.19}\phantom{ (\textcolor{blue}{\textbf{+15.1\%}})} & \textbf{0.80}\phantom{ (\textcolor{blue}{\textbf{+15.1\%}})} \\
DA3-B\textbf{-TCO} & 0.22 (\textbf{+1.0\%}) & 0.83 (\textcolor{blue}{\textbf{+0.7\%}}) & 0.16 (\textbf{+5.0\%}) & 0.76 (\textbf{+1.4\%}) & 0.20 (\textbf{+0.8\%}) & 0.81 (\textcolor{red}{\textbf{-0.4\%}}) & 0.20 (\textbf{+3.9\%}) & 0.84 (\textbf{+0.6\%}) & \textbf{0.19} (\textbf{+2.5\%}) & \textbf{0.81} (\textbf{+0.6\%}) \\
DA3-B\textbf{-Fr} & 0.21 (\textcolor{red}{\textbf{-2.0\%}}) & 0.82 (\textcolor{red}{\textbf{-0.2\%}}) & 0.18 (\textcolor{blue}{\textbf{+19.9\%}}) & 0.77 (\textcolor{blue}{\textbf{+2.9\%}}) & 0.21 (\textbf{+1.7\%}) & 0.81 (\textcolor{blue}{\textbf{+0.8\%}}) & 0.17 (\textcolor{red}{\textbf{-12.3\%}}) & 0.83 (\textcolor{red}{\textbf{-0.8\%}}) & \textbf{0.19} (\textbf{+0.8\%}) & \textbf{0.81} (\textbf{+0.6\%}) \\
DA3-B\textbf{-Ours} & 0.22 (\textcolor{blue}{\textbf{+1.2\%}}) & 0.83 (\textbf{+0.4\%}) & 0.16 (\textbf{+9.8\%}) & 0.75 (\textbf{+0.9\%}) & 0.21 (\textcolor{blue}{\textbf{+2.2\%}}) & 0.81 (\textbf{+0.2\%}) & 0.29 (\textcolor{blue}{\textbf{+54.8\%}}) & 0.87 (\textcolor{blue}{\textbf{+4.4\%}}) & \textbf{0.22} (\textcolor{blue}{\textbf{+16.6\%}}) & \textbf{0.81} (\textcolor{blue}{\textbf{+1.5\%}}) \\
\hline
DA3-S & 0.15\phantom{ (\textcolor{blue}{\textbf{+15.1\%}})} & 0.78\phantom{ (\textcolor{blue}{\textbf{+15.1\%}})} & 0.09\phantom{ (\textcolor{blue}{\textbf{+15.1\%}})} & 0.62\phantom{ (\textcolor{blue}{\textbf{+15.1\%}})} & 0.09\phantom{ (\textcolor{blue}{\textbf{+15.1\%}})} & 0.68\phantom{ (\textcolor{blue}{\textbf{+15.1\%}})} & 0.09\phantom{ (\textcolor{blue}{\textbf{+15.1\%}})} & 0.75\phantom{ (\textcolor{blue}{\textbf{+15.1\%}})} & \textbf{0.10}\phantom{ (\textcolor{blue}{\textbf{+15.1\%}})} & \textbf{0.71}\phantom{ (\textcolor{blue}{\textbf{+15.1\%}})} \\
DA3-S\textbf{-TCO} & 0.15 (\textcolor{blue}{\textbf{+5.6\%}}) & 0.79 (\textbf{+0.4\%}) & 0.09 (\textbf{+2.4\%}) & 0.62 (\textcolor{red}{\textbf{-0.3\%}}) & 0.08 (\textcolor{red}{\textbf{-9.0\%}}) & 0.66 (\textcolor{red}{\textbf{-3.1\%}}) & 0.10 (\textbf{+4.4\%}) & 0.76 (\textbf{+0.9\%}) & \textbf{0.11} (\textbf{+1.5\%}) & \textbf{0.71} (\textcolor{red}{\textbf{-0.5\%}}) \\
DA3-S\textbf{-Fr} & 0.15 (\textbf{+2.4\%}) & 0.78 (\textbf{+0.0\%}) & 0.10 (\textcolor{blue}{\textbf{+19.5\%}}) & 0.66 (\textcolor{blue}{\textbf{+6.4\%}}) & 0.09 (\textcolor{blue}{\textbf{+1.9\%}}) & 0.70 (\textcolor{blue}{\textbf{+1.9\%}}) & 0.09 (\textcolor{red}{\textbf{-5.6\%}}) & 0.74 (\textcolor{red}{\textbf{-1.3\%}}) & \textbf{0.11} (\textbf{+4.0\%}) & \textbf{0.72} (\textbf{+1.5\%}) \\
DA3-S\textbf{-Ours} & 0.14 (\textcolor{red}{\textbf{-2.4\%}}) & 0.79 (\textcolor{blue}{\textbf{+1.0\%}}) & 0.09 (\textbf{+8.3\%}) & 0.64 (\textbf{+3.6\%}) & 0.09 (\textbf{+0.0\%}) & 0.69 (\textbf{+0.3\%}) & 0.16 (\textcolor{blue}{\textbf{+64.8\%}}) & 0.79 (\textcolor{blue}{\textbf{+5.6\%}}) & \textbf{0.12} (\textcolor{blue}{\textbf{+15.6\%}}) & \textbf{0.73} (\textcolor{blue}{\textbf{+2.6\%}}) \\
\hline
\end{tabular}}}
\end{table*}

\begin{table*}[!t]
\centering
\caption{Geometry Estimation results: F1-score ($\uparrow$) under unposed (w/o p.) and posed (w/ p.) modes on 7Scenes~\cite{shotton2013scenes}, ETH3D~\cite{schops2017eth3d}, ScanNet++~\cite{yeshwanth2023scannetpp}, HiRoom~\cite{lin2025da3}, and Mean. Row notation: DA3-G/L/B/S denote DA3-Giant/Large/Base/Small~\cite{lin2025da3}; each pretrained model (e.g., VGGT, $\pi^3$, DA3-G) appears in four consecutive rows --- the baseline and three adapted variants with suffix ``-TCO'' (\textit{TCO}~\cite{tco}), ``-Fr'' (\textit{Free-Geometry}~\cite{dai2026freegeometry}), and ``-Ours'' (our proposed \textit{Self-Geometry}). $\Delta$\% vs.\ baseline in parentheses. Per column, \textcolor{blue}{blue}/\textcolor{red}{red} marks the largest gain/degradation among the three adaptation methods.}
\label{tab:geometry_main}
{%
\resizebox{\textwidth}{!}{
\begin{tabular}{l|ll|ll|ll|ll|ll}
\hline
\multirow{2}{*}{Method} & \multicolumn{2}{c|}{7Scenes} & \multicolumn{2}{c|}{ETH3D} & \multicolumn{2}{c|}{ScanNet++} & \multicolumn{2}{c|}{HiRoom} & \multicolumn{2}{c}{\textcolor{skyblue}{Mean}} \\
 & w/o p. $\uparrow$ & w/ p. $\uparrow$ & w/o p. $\uparrow$ & w/ p. $\uparrow$ & w/o p. $\uparrow$ & w/ p. $\uparrow$ & w/o p. $\uparrow$ & w/ p. $\uparrow$ & w/o p. $\uparrow$ & w/ p. $\uparrow$ \\
\hline
VGGT & 0.43\phantom{ (\textcolor{blue}{\textbf{+15.1\%}})} & 0.37\phantom{ (\textcolor{blue}{\textbf{+15.1\%}})} & 0.52\phantom{ (\textcolor{blue}{\textbf{+15.1\%}})} & 0.43\phantom{ (\textcolor{blue}{\textbf{+15.1\%}})} & 0.60\phantom{ (\textcolor{blue}{\textbf{+15.1\%}})} & 0.40\phantom{ (\textcolor{blue}{\textbf{+15.1\%}})} & 0.60\phantom{ (\textcolor{blue}{\textbf{+15.1\%}})} & 0.68\phantom{ (\textcolor{blue}{\textbf{+15.1\%}})} & \textbf{0.54}\phantom{ (\textcolor{blue}{\textbf{+15.1\%}})} & \textbf{0.47}\phantom{ (\textcolor{blue}{\textbf{+15.1\%}})} \\
VGGT\textbf{-TCO} & 0.44 (\textbf{+3.1\%}) & 0.39 (\textcolor{blue}{\textbf{+4.9\%}}) & 0.49 (\textcolor{red}{\textbf{-4.6\%}}) & 0.40 (\textcolor{red}{\textbf{-7.9\%}}) & 0.48 (\textcolor{red}{\textbf{-19.8\%}}) & 0.36 (\textcolor{red}{\textbf{-10.1\%}}) & 0.44 (\textcolor{red}{\textbf{-26.5\%}}) & 0.51 (\textcolor{red}{\textbf{-24.8\%}}) & \textbf{0.46} (\textcolor{red}{\textbf{-13.4\%}}) & \textbf{0.41} (\textcolor{red}{\textbf{-12.0\%}}) \\
VGGT\textbf{-Fr} & 0.43 (\textbf{+0.4\%}) & 0.37 (\textbf{+1.1\%}) & 0.50 (\textbf{-4.3\%}) & 0.44 (\textbf{+1.1\%}) & 0.59 (\textbf{-1.0\%}) & 0.41 (\textbf{+1.2\%}) & 0.63 (\textcolor{blue}{\textbf{+6.0\%}}) & 0.67 (\textbf{-0.6\%}) & \textbf{0.54} (\textcolor{blue}{\textbf{+0.4\%}}) & \textbf{0.47} (\textbf{+0.5\%}) \\
VGGT\textbf{-Ours} & 0.44 (\textcolor{blue}{\textbf{+3.3\%}}) & 0.38 (\textbf{+4.6\%}) & 0.60 (\textcolor{blue}{\textbf{+16.5\%}}) & 0.47 (\textcolor{blue}{\textbf{+9.3\%}}) & 0.54 (\textbf{-10.6\%}) & 0.41 (\textcolor{blue}{\textbf{+2.5\%}}) & 0.55 (\textbf{-7.5\%}) & 0.64 (\textbf{-4.9\%}) & \textbf{0.53} (\textbf{-0.4\%}) & \textbf{0.48} (\textcolor{blue}{\textbf{+1.8\%}}) \\
\hline
$\pi^3$ & 0.42\phantom{ (\textcolor{blue}{\textbf{+15.1\%}})} & 0.58\phantom{ (\textcolor{blue}{\textbf{+15.1\%}})} & 0.70\phantom{ (\textcolor{blue}{\textbf{+15.1\%}})} & 0.79\phantom{ (\textcolor{blue}{\textbf{+15.1\%}})} & 0.63\phantom{ (\textcolor{blue}{\textbf{+15.1\%}})} & 0.77\phantom{ (\textcolor{blue}{\textbf{+15.1\%}})} & 0.74\phantom{ (\textcolor{blue}{\textbf{+15.1\%}})} & 0.83\phantom{ (\textcolor{blue}{\textbf{+15.1\%}})} & \textbf{0.62}\phantom{ (\textcolor{blue}{\textbf{+15.1\%}})} & \textbf{0.74}\phantom{ (\textcolor{blue}{\textbf{+15.1\%}})} \\
$\pi^3$\textbf{-TCO} & 0.17 (\textcolor{red}{\textbf{-60.1\%}}) & 0.15 (\textcolor{red}{\textbf{-75.0\%}}) & 0.48 (\textcolor{red}{\textbf{-31.5\%}}) & 0.44 (\textcolor{red}{\textbf{-44.3\%}}) & 0.24 (\textcolor{red}{\textbf{-62.1\%}}) & 0.34 (\textcolor{red}{\textbf{-55.8\%}}) & 0.11 (\textcolor{red}{\textbf{-84.9\%}}) & 0.15 (\textcolor{red}{\textbf{-81.6\%}}) & \textbf{0.25} (\textcolor{red}{\textbf{-60.0\%}}) & \textbf{0.27} (\textcolor{red}{\textbf{-63.7\%}}) \\
$\pi^3$\textbf{-Fr} & 0.43 (\textbf{+3.1\%}) & 0.58 (\textcolor{blue}{\textbf{+0.5\%}}) & 0.71 (\textbf{+1.0\%}) & 0.80 (\textbf{+0.2\%}) & 0.51 (\textbf{-19.6\%}) & 0.77 (\textbf{+0.1\%}) & 0.62 (\textbf{-17.1\%}) & 0.81 (\textbf{-2.6\%}) & \textbf{0.56} (\textbf{-9.3\%}) & \textbf{0.74} (\textbf{-0.6\%}) \\
$\pi^3$\textbf{-Ours} & 0.46 (\textcolor{blue}{\textbf{+10.5\%}}) & 0.57 (\textbf{-1.0\%}) & 0.74 (\textcolor{blue}{\textbf{+5.9\%}}) & 0.81 (\textcolor{blue}{\textbf{+2.6\%}}) & 0.65 (\textcolor{blue}{\textbf{+2.2\%}}) & 0.79 (\textcolor{blue}{\textbf{+3.4\%}}) & 0.77 (\textcolor{blue}{\textbf{+3.8\%}}) & 0.84 (\textcolor{blue}{\textbf{+1.3\%}}) & \textbf{0.65} (\textcolor{blue}{\textbf{+5.1\%}}) & \textbf{0.76} (\textcolor{blue}{\textbf{+1.7\%}}) \\
\hline
DA3-G & 0.49\phantom{ (\textcolor{blue}{\textbf{+15.1\%}})} & 0.56\phantom{ (\textcolor{blue}{\textbf{+15.1\%}})} & 0.79\phantom{ (\textcolor{blue}{\textbf{+15.1\%}})} & 0.87\phantom{ (\textcolor{blue}{\textbf{+15.1\%}})} & 0.78\phantom{ (\textcolor{blue}{\textbf{+15.1\%}})} & 0.80\phantom{ (\textcolor{blue}{\textbf{+15.1\%}})} & 0.86\phantom{ (\textcolor{blue}{\textbf{+15.1\%}})} & 0.95\phantom{ (\textcolor{blue}{\textbf{+15.1\%}})} & \textbf{0.73}\phantom{ (\textcolor{blue}{\textbf{+15.1\%}})} & \textbf{0.80}\phantom{ (\textcolor{blue}{\textbf{+15.1\%}})} \\
DA3-G\textbf{-TCO} & 0.51 (\textcolor{blue}{\textbf{+3.1\%}}) & 0.59 (\textcolor{blue}{\textbf{+5.5\%}}) & 0.79 (\textbf{+0.3\%}) & 0.87 (\textbf{+0.1\%}) & 0.76 (\textcolor{red}{\textbf{-2.8\%}}) & 0.80 (\textcolor{red}{\textbf{-0.1\%}}) & 0.84 (\textcolor{red}{\textbf{-2.0\%}}) & 0.94 (\textbf{-1.3\%}) & \textbf{0.73} (\textcolor{red}{\textbf{-0.7\%}}) & \textbf{0.80} (\textcolor{blue}{\textbf{+0.6\%}}) \\
DA3-G\textbf{-Fr} & 0.51 (\textbf{+2.2\%}) & 0.56 (\textcolor{red}{\textbf{-0.9\%}}) & 0.79 (\textcolor{blue}{\textbf{+0.6\%}}) & 0.87 (\textcolor{red}{\textbf{-0.7\%}}) & 0.78 (\textbf{-0.1\%}) & 0.80 (\textbf{-0.0\%}) & 0.86 (\textbf{+0.7\%}) & 0.94 (\textcolor{red}{\textbf{-1.4\%}}) & \textbf{0.74} (\textbf{+0.7\%}) & \textbf{0.79} (\textcolor{red}{\textbf{-0.8\%}}) \\
DA3-G\textbf{-Ours} & 0.51 (\textbf{+2.8\%}) & 0.57 (\textbf{+1.5\%}) & 0.78 (\textcolor{red}{\textbf{-0.2\%}}) & 0.87 (\textcolor{blue}{\textbf{+0.1\%}}) & 0.79 (\textcolor{blue}{\textbf{+0.4\%}}) & 0.81 (\textcolor{blue}{\textbf{+0.5\%}}) & 0.87 (\textcolor{blue}{\textbf{+1.2\%}}) & 0.95 (\textbf{-0.5\%}) & \textbf{0.74} (\textcolor{blue}{\textbf{+0.9\%}}) & \textbf{0.80} (\textbf{+0.3\%}) \\
\hline
DA3-L & 0.51\phantom{ (\textcolor{blue}{\textbf{+15.1\%}})} & 0.48\phantom{ (\textcolor{blue}{\textbf{+15.1\%}})} & 0.69\phantom{ (\textcolor{blue}{\textbf{+15.1\%}})} & 0.75\phantom{ (\textcolor{blue}{\textbf{+15.1\%}})} & 0.69\phantom{ (\textcolor{blue}{\textbf{+15.1\%}})} & 0.76\phantom{ (\textcolor{blue}{\textbf{+15.1\%}})} & 0.69\phantom{ (\textcolor{blue}{\textbf{+15.1\%}})} & 0.88\phantom{ (\textcolor{blue}{\textbf{+15.1\%}})} & \textbf{0.65}\phantom{ (\textcolor{blue}{\textbf{+15.1\%}})} & \textbf{0.72}\phantom{ (\textcolor{blue}{\textbf{+15.1\%}})} \\
DA3-L\textbf{-TCO} & 0.51 (\textcolor{red}{\textbf{-0.7\%}}) & 0.48 (\textcolor{blue}{\textbf{+1.3\%}}) & 0.68 (\textbf{-1.4\%}) & 0.75 (\textcolor{red}{\textbf{-0.0\%}}) & 0.65 (\textcolor{red}{\textbf{-5.5\%}}) & 0.76 (\textcolor{red}{\textbf{-0.9\%}}) & 0.74 (\textcolor{blue}{\textbf{+7.5\%}}) & 0.83 (\textbf{-5.2\%}) & \textbf{0.65} (\textbf{+0.0\%}) & \textbf{0.71} (\textbf{-1.6\%}) \\
DA3-L\textbf{-Fr} & 0.53 (\textcolor{blue}{\textbf{+3.5\%}}) & 0.48 (\textbf{-0.5\%}) & 0.68 (\textcolor{red}{\textbf{-2.7\%}}) & 0.76 (\textcolor{blue}{\textbf{+0.8\%}}) & 0.69 (\textcolor{blue}{\textbf{+0.6\%}}) & 0.76 (\textbf{+0.0\%}) & 0.57 (\textcolor{red}{\textbf{-17.3\%}}) & 0.77 (\textcolor{red}{\textbf{-12.5\%}}) & \textbf{0.62} (\textcolor{red}{\textbf{-4.5\%}}) & \textbf{0.69} (\textcolor{red}{\textbf{-3.7\%}}) \\
DA3-L\textbf{-Ours} & 0.51 (\textbf{+0.4\%}) & 0.47 (\textcolor{red}{\textbf{-1.3\%}}) & 0.69 (\textbf{-1.1\%}) & 0.75 (\textbf{+0.1\%}) & 0.69 (\textbf{+0.5\%}) & 0.76 (\textcolor{blue}{\textbf{+0.1\%}}) & 0.72 (\textbf{+4.3\%}) & 0.88 (\textbf{-0.1\%}) & \textbf{0.65} (\textcolor{blue}{\textbf{+1.1\%}}) & \textbf{0.72} (\textbf{-0.2\%}) \\
\hline
DA3-B & 0.50\phantom{ (\textcolor{blue}{\textbf{+15.1\%}})} & 0.50\phantom{ (\textcolor{blue}{\textbf{+15.1\%}})} & 0.49\phantom{ (\textcolor{blue}{\textbf{+15.1\%}})} & 0.66\phantom{ (\textcolor{blue}{\textbf{+15.1\%}})} & 0.48\phantom{ (\textcolor{blue}{\textbf{+15.1\%}})} & 0.67\phantom{ (\textcolor{blue}{\textbf{+15.1\%}})} & 0.23\phantom{ (\textcolor{blue}{\textbf{+15.1\%}})} & 0.72\phantom{ (\textcolor{blue}{\textbf{+15.1\%}})} & \textbf{0.42}\phantom{ (\textcolor{blue}{\textbf{+15.1\%}})} & \textbf{0.64}\phantom{ (\textcolor{blue}{\textbf{+15.1\%}})} \\
DA3-B\textbf{-TCO} & 0.48 (\textcolor{red}{\textbf{-3.8\%}}) & 0.51 (\textcolor{blue}{\textbf{+2.6\%}}) & 0.49 (\textcolor{red}{\textbf{-0.2\%}}) & 0.68 (\textcolor{blue}{\textbf{+2.7\%}}) & 0.44 (\textcolor{red}{\textbf{-7.5\%}}) & 0.67 (\textcolor{red}{\textbf{-0.1\%}}) & 0.26 (\textbf{+11.3\%}) & 0.65 (\textcolor{red}{\textbf{-9.7\%}}) & \textbf{0.42} (\textbf{-1.8\%}) & \textbf{0.63} (\textcolor{red}{\textbf{-1.5\%}}) \\
DA3-B\textbf{-Fr} & 0.49 (\textbf{-2.0\%}) & 0.50 (\textbf{-0.5\%}) & 0.50 (\textbf{+1.3\%}) & 0.66 (\textcolor{red}{\textbf{-0.2\%}}) & 0.48 (\textcolor{blue}{\textbf{+1.2\%}}) & 0.68 (\textcolor{blue}{\textbf{+0.5\%}}) & 0.19 (\textcolor{red}{\textbf{-18.4\%}}) & 0.68 (\textbf{-5.4\%}) & \textbf{0.41} (\textcolor{red}{\textbf{-2.4\%}}) & \textbf{0.63} (\textbf{-1.5\%}) \\
DA3-B\textbf{-Ours} & 0.50 (\textcolor{blue}{\textbf{+0.2\%}}) & 0.50 (\textcolor{red}{\textbf{-0.8\%}}) & 0.50 (\textcolor{blue}{\textbf{+1.9\%}}) & 0.67 (\textbf{+0.8\%}) & 0.48 (\textbf{+0.2\%}) & 0.67 (\textbf{+0.3\%}) & 0.43 (\textcolor{blue}{\textbf{+85.2\%}}) & 0.72 (\textcolor{blue}{\textbf{+0.4\%}}) & \textbf{0.48} (\textcolor{blue}{\textbf{+12.2\%}}) & \textbf{0.64} (\textcolor{blue}{\textbf{+0.3\%}}) \\
\hline
DA3-S & 0.39\phantom{ (\textcolor{blue}{\textbf{+15.1\%}})} & 0.47\phantom{ (\textcolor{blue}{\textbf{+15.1\%}})} & 0.37\phantom{ (\textcolor{blue}{\textbf{+15.1\%}})} & 0.64\phantom{ (\textcolor{blue}{\textbf{+15.1\%}})} & 0.33\phantom{ (\textcolor{blue}{\textbf{+15.1\%}})} & 0.53\phantom{ (\textcolor{blue}{\textbf{+15.1\%}})} & 0.18\phantom{ (\textcolor{blue}{\textbf{+15.1\%}})} & 0.51\phantom{ (\textcolor{blue}{\textbf{+15.1\%}})} & \textbf{0.32}\phantom{ (\textcolor{blue}{\textbf{+15.1\%}})} & \textbf{0.54}\phantom{ (\textcolor{blue}{\textbf{+15.1\%}})} \\
DA3-S\textbf{-TCO} & 0.39 (\textbf{-1.1\%}) & 0.47 (\textbf{+0.7\%}) & 0.38 (\textbf{+2.0\%}) & 0.66 (\textcolor{blue}{\textbf{+3.1\%}}) & 0.30 (\textcolor{red}{\textbf{-7.9\%}}) & 0.53 (\textcolor{red}{\textbf{-0.9\%}}) & 0.21 (\textbf{+13.8\%}) & 0.42 (\textcolor{red}{\textbf{-17.6\%}}) & \textbf{0.32} (\textbf{+0.2\%}) & \textbf{0.52} (\textcolor{red}{\textbf{-3.4\%}}) \\
DA3-S\textbf{-Fr} & 0.41 (\textcolor{blue}{\textbf{+4.5\%}}) & 0.47 (\textcolor{red}{\textbf{-0.1\%}}) & 0.49 (\textcolor{blue}{\textbf{+31.5\%}}) & 0.64 (\textbf{+1.3\%}) & 0.33 (\textbf{+0.8\%}) & 0.53 (\textbf{-0.2\%}) & 0.16 (\textcolor{red}{\textbf{-12.1\%}}) & 0.51 (\textbf{-0.8\%}) & \textbf{0.35} (\textbf{+9.1\%}) & \textbf{0.54} (\textbf{+0.1\%}) \\
DA3-S\textbf{-Ours} & 0.38 (\textcolor{red}{\textbf{-2.1\%}}) & 0.47 (\textcolor{blue}{\textbf{+1.4\%}}) & 0.41 (\textbf{+11.0\%}) & 0.65 (\textbf{+2.1\%}) & 0.34 (\textcolor{blue}{\textbf{+4.4\%}}) & 0.53 (\textcolor{blue}{\textbf{+0.4\%}}) & 0.31 (\textcolor{blue}{\textbf{+70.4\%}}) & 0.55 (\textcolor{blue}{\textbf{+7.0\%}}) & \textbf{0.36} (\textcolor{blue}{\textbf{+13.9\%}}) & \textbf{0.55} (\textcolor{blue}{\textbf{+2.7\%}}) \\
\hline
\end{tabular}}}
\end{table*}

\begin{figure*}[t]
\centering
\includegraphics[width=\textwidth]{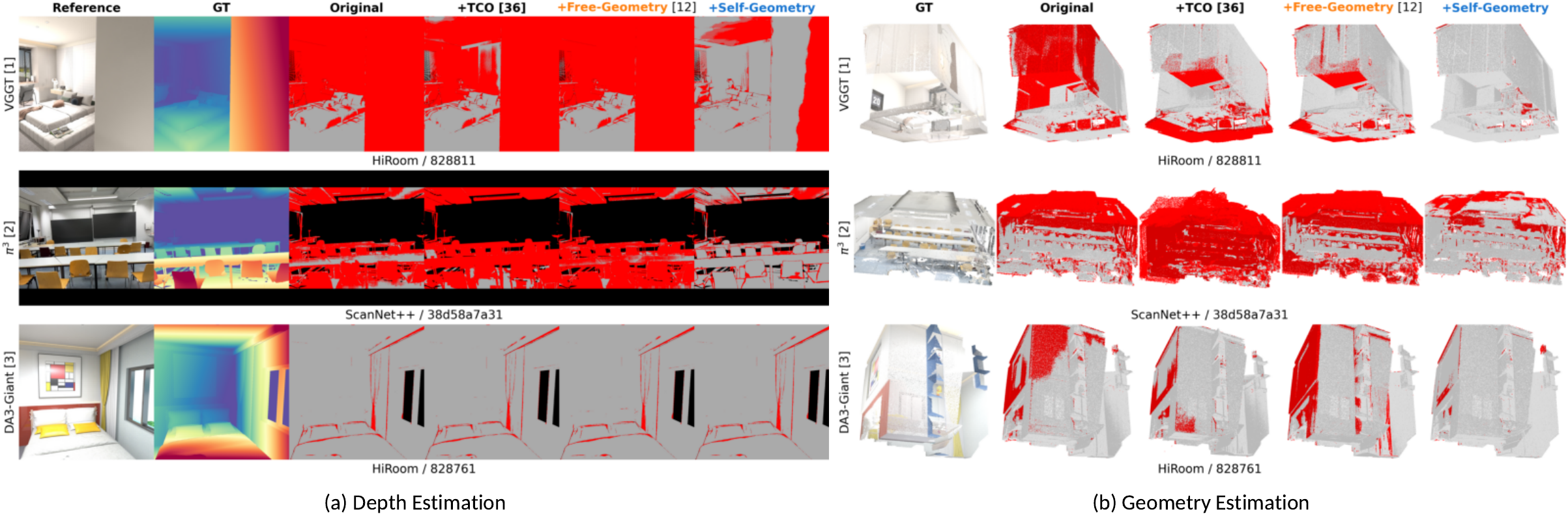}
\caption{Qualitative comparison of pretrained \textit{VFMs} against our proposed \textit{Self-Geometry} across representative scenes (rows: \textit{VGGT}~\cite{wang2025vggt}, $\pi^3$~\cite{yang2025pi3}, \textit{DA3-Giant}~\cite{lin2025da3}; scene identifier below each row). \textbf{(a) Depth Estimation}: from left to right, Reference RGB, GT depth, and depth error maps for Original, +TCO~\cite{tco}, \textcolor{orange}{+Free-Geometry~\cite{dai2026freegeometry}}, and \textcolor{blue}{+Self-Geometry} (Ours). \textbf{(b) Geometry Estimation}: from left to right, GT fused pointcloud and geometry error maps for the same four methods. \textcolor{red}{Red} regions denote errors exceeding the benchmark threshold; \textcolor{gray}{gray} regions denote correctly reconstructed pixels. \textit{Best viewed in zoom.}}
\label{fig:qual}
\end{figure*}

{\color{black}
We evaluate our proposed \textit{Self-Geometry} on Pose Estimation and Geometry Estimation across multiple pretrained \textit{Vision Foundation Models (VFMs)} (VGGT~\cite{wang2025vggt}, $\pi^3$~\cite{yang2025pi3}, and DA3~\cite{lin2025da3}) and four benchmark datasets (7Scenes~\cite{shotton2013scenes}, ETH3D~\cite{schops2017eth3d}, ScanNet++~\cite{yeshwanth2023scannetpp}, and HiRoom~\cite{lin2025da3}), compare it against \textit{Free-Geometry}~\cite{dai2026freegeometry} and \textit{TCO}~\cite{tco}, and analyze the effectiveness of the proposed \textit{Epipolar Consistency Loss (EC Loss)} (Sec.~\ref{para:ec}), \textit{Multi-View Consistency Loss (MVC Loss)} (Sec.~\ref{para:mvc}), \textit{Geometric Disentanglement Optimization (GDO)} (Sec.~\ref{subsec:gdo}), and \textit{Frame Angular-Neighbor (FAN)} (Sec.~\ref{subsec:fan}) through qualitative comparisons, ablation studies, and complexity analysis. Additional implementation details are provided in Sec.~\ref{sec:supp_training}.

\subsection{Experimental Setup}
\label{subsec:experimental_setup}

\textbf{Datasets.} Following the DA3 benchmark protocol, including its per-scene sampling of up to 100 frames, we evaluate our proposed \textit{Self-Geometry} on 7Scenes~\cite{shotton2013scenes}, ETH3D~\cite{schops2017eth3d}, ScanNet++~\cite{yeshwanth2023scannetpp}, and HiRoom~\cite{lin2025da3}. Since our proposed \textit{Self-Geometry} targets pose and geometry estimation in indoor and outdoor scenes, we exclude the object-centric DTU~\cite{jensen2014dtu}.

\textbf{Pose Estimation.}\label{para:pose_estimation} Following DA3~\cite{lin2025da3}, we report AUC@1, AUC@3, and AUC@30 (thresholds in degrees) of pairwise pose error, averaged across benchmark scenes.

\textbf{Geometry Estimation.}\label{para:geometry_estimation} Following DA3~\cite{lin2025da3}, we report F1-score under posed (GT poses; depth only) and unposed (predicted poses; pose and depth) modes.

\textbf{Baselines.} To evaluate generalizability across \textit{VFM} architectures and scales, we experiment on three feed-forward \textit{VFMs} (VGGT~\cite{wang2025vggt}, $\pi^3$~\cite{yang2025pi3}, and DA3~\cite{lin2025da3} with Giant, Large, Base, and Small variants). For each model-dataset pair, we compare the frozen pretrained \textit{VFM} to the same model adapted by our proposed \textit{Self-Geometry}. We additionally include the two most recent comparison baselines, \textit{Free-Geometry}~\cite{dai2026freegeometry} and \textit{TCO}~\cite{tco}. For \textit{TCO}, since GT camera poses are unavailable, we use the pretrained \textit{VFM}'s predicted camera poses as its auxiliary prior. \textit{For fair comparison}, both baselines share the same DA3 evaluation protocol as our proposed \textit{Self-Geometry}, and adopt the training configurations of their official implementations. For \textit{VFMs} unsupported by these implementations, we re-implement and re-train each baseline.

\textbf{Training Details.} We instantiate the proposed \textit{Lightweight TTA} (Sec.~\ref{subsec:lightweight_tta}) with \textit{LoRA}~\cite{hu2022lora}; the pretrained parameters remain frozen, and each \textit{VFM} is adapted per-scene with identical optimization settings. The total loss combines the two primary losses (Eqs.~\eqref{eq:mvc} and \eqref{eq:ec}) with three self-supervised auxiliary regularizers that suppress excessive drift from the pretrained \textit{VFM}, formalized in Sec.~\ref{sec:supp_total_loss}. We optimize each scene with AdamW~\cite{loshchilov2019adamw} for 50 \textit{TTA} iterations, selecting the checkpoint minimizing the scale-invariant $\sqrt{\mathcal{L}_{\mathrm{ec}} \cdot \mathcal{L}_{\mathrm{mvc}}}$ (Eqs.~\eqref{eq:ec} and \eqref{eq:mvc}). Training details are provided in Sec.~\ref{sec:supp_training}.

\subsection{Quantitative Results}
\label{subsec:quantitative_results}

Tab.~\ref{tab:pose_main} and Tab.~\ref{tab:geometry_main} present pose and geometry estimation results, respectively. Across six pretrained \textit{VFMs}, our proposed \textit{Self-Geometry} outperforms \textit{Free-Geometry}~\cite{dai2026freegeometry} and \textit{TCO}~\cite{tco} on most Mean columns and achieves the largest Mean improvements over the frozen baselines in both tasks.

On $\pi^3$~\cite{yang2025pi3}, \textit{TCO}~\cite{tco} collapses across all four datasets, degrading Mean pose AUC@3 by \textbf{-91.5\%} and Mean geometry w/o p.\ by \textbf{-60.0\%} (Tab.~\ref{tab:pose_main} and Tab.~\ref{tab:geometry_main}). In our GT-free setting where \textit{TCO}'s pose prior is derived from the pretrained \textit{VFM}'s own prediction (Sec.~\ref{subsec:experimental_setup}), \textit{TCO}'s optimization diverges on $\pi^3$. Our proposed \textit{Self-Geometry}, by contrast, delivers consistent Mean gains on $\pi^3$ (AUC@3 \textbf{+8.3\%}, geometry w/o p.\ \textbf{+5.1\%}), highlighting its robustness across pretrained \textit{VFMs}.

For Pose Estimation, relative gains grow as the AUC threshold tightens. On ETH3D~\cite{schops2017eth3d}, which comprises wide-baseline outdoor scenes, our proposed \textit{Self-Geometry} improves the AUC@30 / AUC@3 of VGGT~\cite{wang2025vggt} by \textbf{+9.2\% / +37.3\%} and of $\pi^3$~\cite{yang2025pi3} by \textbf{+5.0\% / +25.1\%}, as the wide baselines strengthen correspondence-based geometric constraints. On HiRoom~\cite{lin2025da3}, a challenging synthetic benchmark with varied indoor illumination and abundant fine structures, the gains are most pronounced for the smaller pretrained \textit{VFMs}: DA3-Base and DA3-Small improve geometry w/o p.\ by \textbf{+85.2\%} and \textbf{+70.4\%}, respectively, suggesting that our proposed \textit{Self-Geometry} is particularly effective in the under-fitted regime.

On ScanNet++~\cite{yeshwanth2023scannetpp} and HiRoom~\cite{lin2025da3}, VGGT+Ours exhibits relative degradations in a subset of columns (e.g., AUC@3 \textbf{-2.9\% / -8.2\%}, geometry w/o p.\ \textbf{-10.6\% / -7.5\%}). These relative degradations correspond to small absolute drops within 6\,pp, and the Mean pose improvements of VGGT+Ours remain positive (AUC@3 Mean \textbf{+3.3\%}, AUC@30 Mean \textbf{+1.4\%}).

\subsection{Qualitative Results}
\label{subsec:qualitative_results}

We qualitatively compare diverse pretrained \textit{VFMs} such as VGGT~\cite{wang2025vggt}, $\pi^3$~\cite{yang2025pi3}, and DA3-Giant~\cite{lin2025da3} against \textit{Free-Geometry}~\cite{dai2026freegeometry} and our proposed \textit{Self-Geometry} on representative scenes from HiRoom~\cite{lin2025da3} and ScanNet++~\cite{yeshwanth2023scannetpp}, along two axes: depth (Fig.~\ref{fig:qual}(a)) and 3D geometry (Fig.~\ref{fig:qual}(b)).

\textbf{Depth Estimation Results.} Fig.~\ref{fig:qual}(a) visualizes per-pixel depth error against GT. Original, \textit{TCO}~\cite{tco}, and \textit{Free-Geometry}~\cite{dai2026freegeometry} all leave widespread \textcolor{red}{red} regions, and \textit{TCO} expands them further on $\pi^3$~\cite{yang2025pi3}. In contrast, our proposed \textit{Self-Geometry} substantially reduces them across all three backbones, consistent with the F1-score gains in Tab.~\ref{tab:geometry_main}.

\textbf{Geometry Estimation Results.} Fig.~\ref{fig:qual}(b) shows fused pointclouds. Compared to Original, \textit{TCO}~\cite{tco}, and \textit{Free-Geometry}~\cite{dai2026freegeometry}, which exhibit scattered \textcolor{red}{red} outliers (most severe for \textit{TCO} on $\pi^3$~\cite{yang2025pi3}), our proposed \textit{Self-Geometry} yields noticeably cleaner surface reconstructions with fewer error regions. Additional qualitative results across more scenes are provided in Figs.~\ref{fig:supp_qual_7scenes}, \ref{fig:supp_qual_scannetpp}, and \ref{fig:supp_qual_hiroom}.}

\section{Ablation Study}
\label{sec:ablation}

This section isolates the contribution from each of the three core components in our proposed \textit{Self-Geometry} (\textit{GDO} (Sec.~\ref{subsec:gdo}), \textit{Pseudo-Correspondence Filtering} (Sec.~\ref{para:filtering}), and \textit{FAN} (Sec.~\ref{subsec:fan})), and profiles its computational complexity (Sec.~\ref{subsec:complexity}). Component ablations are conducted on VGGT~\cite{wang2025vggt} with ETH3D~\cite{schops2017eth3d}.

\begin{table*}[t]
\centering
\sbox{\gdoTableBox}{%
\resizebox{0.62\textwidth}{!}{%
\begin{tabular}{l|l|ccc|cc}
\hline
\multicolumn{2}{c|}{Variant} & AUC@1 $\uparrow$ & AUC@3 $\uparrow$ & AUC@30 $\uparrow$ & w/ p. $\uparrow$ & w/o p. $\uparrow$ \\
\hline
-        & Baseline                                    & 0.03          & 0.20          & 0.76          & 0.43          & 0.52 \\
\hline
\multicolumn{7}{l}{\textit{(i) Loss Components}} \\
\hline
A        & w/o $\mathcal{L}_{\text{mvc}}$              & 0.04          & 0.20          & 0.74          & 0.45          & 0.49 \\
B        & w/o $\mathcal{L}_{\text{ec}}$               & 0.05          & 0.23          & 0.82          & \textbf{0.50} & 0.59 \\
\hline
\multicolumn{7}{l}{\textit{(ii) Gradient Disentanglement}} \\
\hline
C        & w/o Gradient Disent.                        & 0.05          & 0.27          & 0.82          & 0.45          & 0.59 \\
D        & $\nabla \mathcal{L}_{\mathrm{mvc}}$ Disent. & 0.06          & 0.25          & 0.78          & 0.44          & 0.52 \\
E        & $\nabla \mathcal{L}_{\mathrm{ec}}$ Disent. \textbf{(Ours)} & \textbf{0.07} & \textbf{0.27} & \textbf{0.83} & \textbf{0.47} & \textbf{0.60} \\
F        & Bidirectional Disent.                       & 0.06          & 0.26          & 0.82          & 0.46          & 0.59 \\
\hline
\end{tabular}}%
}%
\begin{minipage}[c]{0.62\textwidth}
\centering
\captionof{table}{Ablations for \textit{GDO} (Sec.~\ref{subsec:gdo}) on VGGT~\cite{wang2025vggt} with ETH3D~\cite{schops2017eth3d}. Block (i) toggles $\mathcal{L}_{\text{ec}}$ (Eq.~\eqref{eq:ec}) and $\mathcal{L}_{\text{mvc}}$ (Eq.~\eqref{eq:mvc}); Block (ii) varies \textit{GD} (Eq.~\eqref{eq:gdo_projection}) strategies. Pose (AUC@1/3/30 $\uparrow$) and Geometry (F1 $\uparrow$; w/ p., w/o p.). \textbf{Bold}: best per column.}
\label{tab:ablation_gdo}
\usebox{\gdoTableBox}
\end{minipage}
\hfill
\begin{minipage}[c]{\dimexpr\textwidth-\wd\gdoTableBox-2em\relax}
\centering
\includegraphics[width=\linewidth]{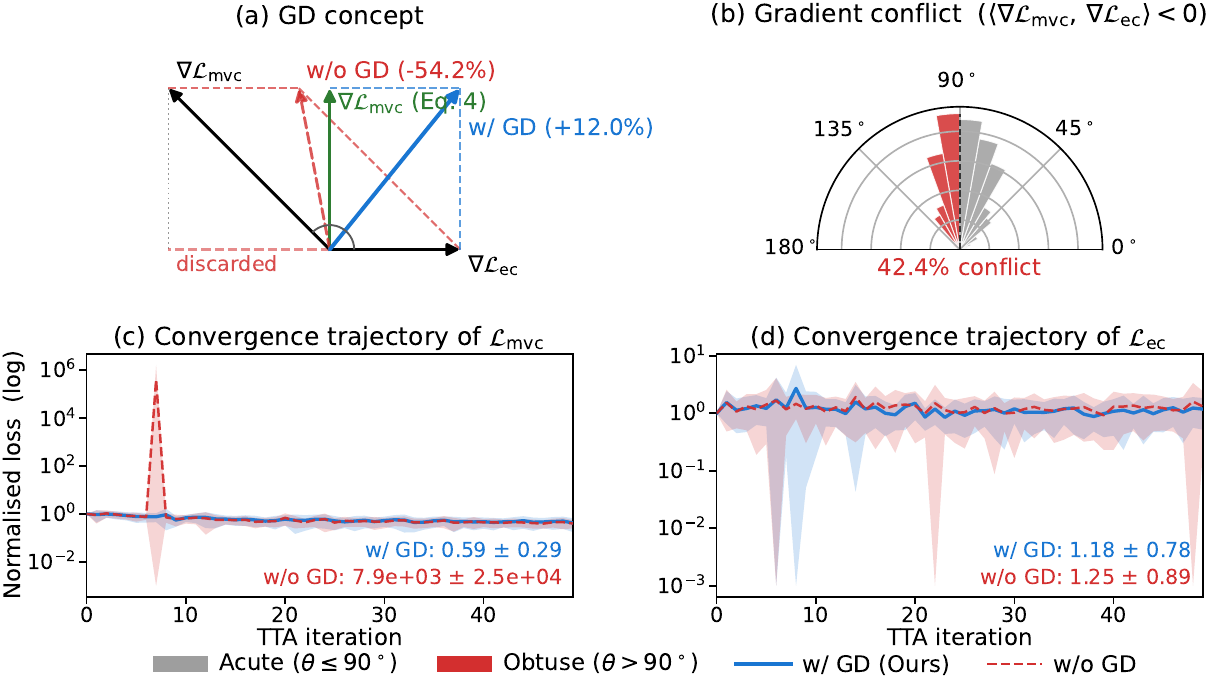}
\captionof{figure}{Visualization of our proposed \textit{GD} (Eq.~\eqref{eq:gdo_projection}) on VGGT~\cite{wang2025vggt} with ETH3D~\cite{schops2017eth3d}. (a) Concept of \textit{GD} on a representative gradient pair. (b) Polar histogram of $\angle(\nabla \mathcal{L}_{\mathrm{ec}}, \nabla \mathcal{L}_{\mathrm{mvc}})$ aggregated over all ETH3D scenes and \textit{TTA} iterations, with obtuse bins in \textcolor{red}{red} and acute bins in \textcolor{gray}{gray}. (c, d) Convergence of $\mathcal{L}_{\mathrm{mvc}}$ and $\mathcal{L}_{\mathrm{ec}}$ across \textit{TTA} iterations, w/ \textit{GD}: \textcolor{blue}{blue}, w/o: \textcolor{red}{red}.}
\label{fig:ablation_gdo}
\end{minipage}
\end{table*}

\begin{table*}[t]
\centering
\caption{Ablations for \textit{Pseudo-Correspondence Filtering} (Sec.~\ref{para:filtering}) on VGGT~\cite{wang2025vggt} with ETH3D~\cite{schops2017eth3d}. Variants A to D toggle the two filtering stages using $\mathcal{L}_{\text{ec}}$ (Eq.~\eqref{eq:ec}) and $\mathcal{L}_{\text{mvc}}$ (Eq.~\eqref{eq:mvc}). Precision/Recall/F1 (vs.\ GT), and downstream Pose (AUC@1/3/30 $\uparrow$) / Geometry (F1 $\uparrow$; w/ p., w/o p.) after \textit{TTA}. \textbf{Bold}: best per column.}
\label{tab:ablation_corr}
\resizebox{\textwidth}{!}{%
\begin{tabular}{l|l|ccc|ccc|cc}
\hline
\multicolumn{2}{c|}{\multirow{2}{*}{Variant}}
 & \multicolumn{3}{c|}{Correspondence Quality}
 & \multicolumn{3}{c|}{Pose Estimation}
 & \multicolumn{2}{c}{Geometry Estimation} \\
\multicolumn{2}{c|}{}
 & Precision $\uparrow$ & Recall $\uparrow$ & F1 $\uparrow$
 & AUC@1 $\uparrow$ & AUC@3 $\uparrow$ & AUC@30 $\uparrow$
 & w/ p. $\uparrow$ & w/o p. $\uparrow$ \\
\hline
-        & Baseline                                & --            & --            & --            & 0.03          & 0.20          & 0.76          & 0.43          & 0.52 \\
A        & Raw (see Fig.~\ref{fig:ablation_corr}(a))                                                                     & 0.39          & \textbf{1.00} & 0.73          & 0.01          & 0.04          & 0.37          & 0.17          & 0.17 \\
B        & $\mathcal{L}_{\text{ec}}$ filter (see Fig.~\ref{fig:ablation_corr}(a))                                        & 0.56          & 0.93          & 0.74          & 0.03          & 0.14          & 0.54          & 0.31          & 0.32 \\
C        & $\mathcal{L}_{\text{mvc}}$ filter                                                                             & 0.62          & 0.93          & \textbf{0.76} & 0.06          & 0.24          & 0.79          & 0.46          & 0.54 \\
D        & $\mathcal{L}_{\text{ec}}$ + $\mathcal{L}_{\text{mvc}}$ filter (see Fig.~\ref{fig:ablation_corr}(a)) \textbf{(Ours)}    & \textbf{0.63} & 0.88          & 0.75          & \textbf{0.07} & \textbf{0.27} & \textbf{0.83} & \textbf{0.47} & \textbf{0.60} \\
\hline
\end{tabular}}
\end{table*}

\begin{figure*}[t]
\centering
\includegraphics[width=\linewidth]{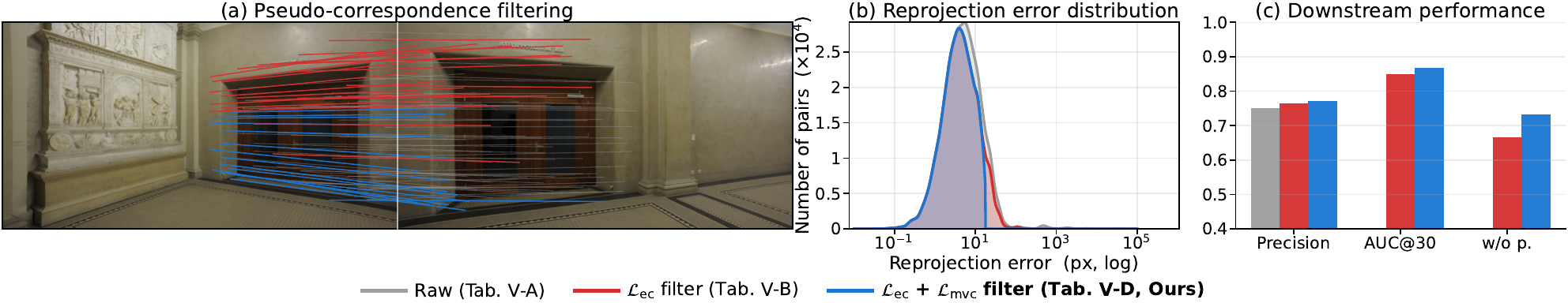}
\caption{Visualization of our proposed \textit{Pseudo-Correspondence Filtering} (Sec.~\ref{para:filtering}) on VGGT~\cite{wang2025vggt} with the ETH3D~\cite{schops2017eth3d} \textit{relief} scene, comparing Raw (\textcolor{gray}{gray}), $\mathcal{L}_{\mathrm{ec}}$ filter (\textcolor{red}{red}), and $\mathcal{L}_{\mathrm{ec}}$ + $\mathcal{L}_{\mathrm{mvc}}$ filter (Ours, \textcolor{blue}{blue}). (a) Filtering on a representative pair. (b) Reprojection error distribution across all pairs. (c) Downstream Precision, pose AUC@30, and geometry F1 (w/o p.).}
\label{fig:ablation_corr}
\end{figure*}

\textbf{Ablations for GDO.}\label{para:ablation_gdo} We ablate the proposed \textit{GDO} (Sec.~\ref{subsec:gdo}) in Tab.~\ref{tab:ablation_gdo} and analyze it in Fig.~\ref{fig:ablation_gdo}. w/o $\mathcal{L}_{\text{mvc}}$ (Tab.~\ref{tab:ablation_gdo}-(i)-A) yields no meaningful improvement over the baseline, and w/o $\mathcal{L}_{\text{ec}}$ (Tab.~\ref{tab:ablation_gdo}-(i)-B) improves over the baseline but underperforms the variants that combine the two losses (Tab.~\ref{tab:ablation_gdo}-(ii)-C, D, E, F), indicating that combining the two losses is essential. However, this combination alone is insufficient: as shown in Fig.~\ref{fig:ablation_gdo}(b), the two losses produce \textit{gradient conflict} in 42.4\% of TTA iterations across all ETH3D scenes, and w/o Gradient Disent.\ (Tab.~\ref{tab:ablation_gdo}-(ii)-C), which combines both losses without \textit{GD}, exhibits a drop in w/ p.\ ($0.50 \to 0.45$) relative to Tab.~\ref{tab:ablation_gdo}-(i)-B. This motivates the need for explicit \textit{GD}. The proposed \textit{GD} direction $\nabla \mathcal{L}_{\text{ec}}$ Disent.\ (Tab.~\ref{tab:ablation_gdo}-(ii)-E, Ours) attains the best performance across most metrics over the remaining GD variants ($\nabla \mathcal{L}_{\text{mvc}}$ Disent.\ (Tab.~\ref{tab:ablation_gdo}-(ii)-D) and Bidirectional Disent.\ (Tab.~\ref{tab:ablation_gdo}-(ii)-F)), validating our \textit{GD} design that lets the two losses supervise camera pose and depth in a disentangled and complementary manner. Additionally, as shown in Fig.~\ref{fig:ablation_gdo}(c, d), the proposed \textit{GD} stabilizes the convergence of both losses.

\textbf{Ablations for Pseudo-Correspondence Filtering.}\label{para:ablation_filtering} We ablate the proposed \textit{Pseudo-Correspondence Filtering} (Sec.~\ref{para:filtering}) in Tab.~\ref{tab:ablation_corr} and analyze it in Fig.~\ref{fig:ablation_corr}. $\mathcal{L}_{\text{ec}}$ + $\mathcal{L}_{\text{mvc}}$ filter (Tab.~\ref{tab:ablation_corr}-D, Ours), the sequential combination that the proposed \textit{Pseudo-Correspondence Filtering} adopts, outperforms $\mathcal{L}_{\text{ec}}$ filter (Tab.~\ref{tab:ablation_corr}-B) on both Precision and downstream metrics, indicating that Precision of pseudo-correspondences is critical for downstream performance. As shown in Fig.~\ref{fig:ablation_corr}(b), Tab.~\ref{tab:ablation_corr}-B and Tab.~\ref{tab:ablation_corr}-D trim the large-error tail of the raw reprojection error distribution, securing Precision. Additionally, as shown in Fig.~\ref{fig:ablation_corr}(a), the two filters remove different types of mismatch, confirming that their combination operates complementarily.

\begin{table*}[t]
\centering
\sbox{\fanTableBox}{%
\resizebox{0.62\textwidth}{!}{%
\begin{tabular}{l|l|ccc|cc}
\hline
\multicolumn{2}{c|}{Variant} & AUC@1 $\uparrow$ & AUC@3 $\uparrow$ & AUC@30 $\uparrow$ & w/ p. $\uparrow$ & w/o p. $\uparrow$ \\
\hline
-        & Baseline                                    & 0.03          & 0.20          & 0.76          & 0.43          & 0.52 \\
\hline
\multicolumn{7}{l}{\textit{(i) SO(3) Bin Width}} \\
\hline
A        & 15$^\circ$ (9 bins) \textbf{(Ours)}                  & \textbf{0.07} & \textbf{0.27} & 0.83          & 0.47          & 0.60 \\
B        & 30$^\circ$ (6 bins)                         & 0.04          & 0.24          & \textbf{0.83} & \textbf{0.48} & 0.62 \\
C        & 45$^\circ$ (4 bins)                         & 0.06          & 0.24          & 0.81          & 0.46          & \textbf{0.62} \\
\hline
\multicolumn{7}{l}{\textit{(ii) Target View Selection}} \\
\hline
A        & $t=0$ (Fixed Target View; see Fig.~\ref{fig:ablation_fan}-(a)) & 0.04          & 0.22          & 0.76          & 0.41          & 0.54 \\
B        & \textit{GRV} (Eq.~\eqref{eq:grv_selection}; see Fig.~\ref{fig:ablation_fan}-(b)) \textbf{(Ours)} & \textbf{0.07} & \textbf{0.27} & \textbf{0.83} & \textbf{0.47} & \textbf{0.60} \\
C        & Dynamic \textit{GRV}                        & 0.07          & 0.26          & 0.83          & 0.46          & 0.58 \\
\hline
\multicolumn{7}{l}{\textit{(iii) Source View Ordering}} \\
\hline
A        & Sequential Order \textbf{(Ours)}                     & \textbf{0.07} & \textbf{0.27} & \textbf{0.83} & \textbf{0.47} & 0.60 \\
B        & Asc. Match Count                            & 0.06          & 0.26          & 0.82          & 0.47          & 0.59 \\
C        & Desc. Match Count                           & 0.07          & 0.26          & 0.82          & 0.45          & \textbf{0.61} \\
\hline
\end{tabular}}%
}%
\begin{minipage}[c]{0.62\textwidth}
\centering
\captionof{table}{Ablations for \textit{FAN} (Sec.~\ref{subsec:fan}) on VGGT~\cite{wang2025vggt} with ETH3D~\cite{schops2017eth3d}. Block (i) sweeps \textit{SO(3) bin width}; Block (ii) varies \textit{target view selection}; Block (iii) varies \textit{source view ordering} within each SO(3) bin. Pose (AUC@1/3/30 $\uparrow$) and Geometry (F1 $\uparrow$; w/ p., w/o p.). \textbf{Bold}: best per column.}
\label{tab:ablation_fan}
\usebox{\fanTableBox}
\end{minipage}
\hfill
\begin{minipage}[c]{\dimexpr\textwidth-\wd\fanTableBox-2em\relax}
\centering
\includegraphics[width=\linewidth]{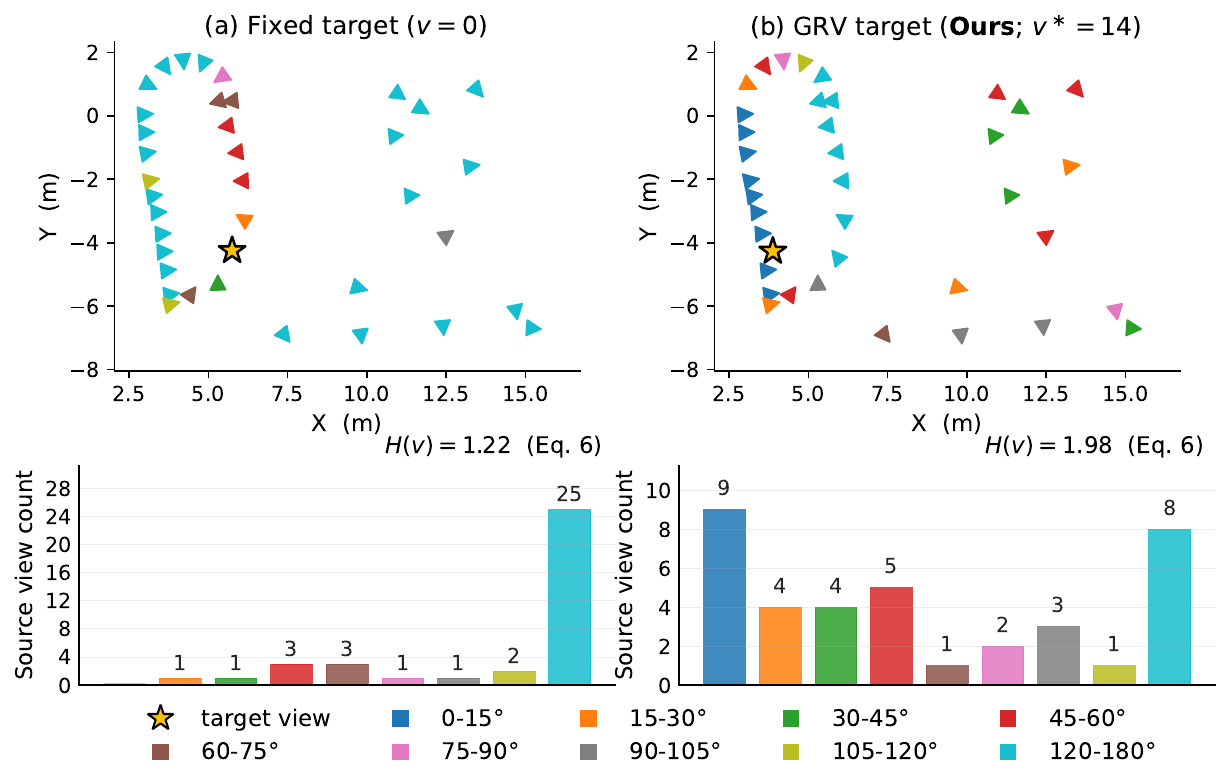}
\captionof{figure}{Comparison of target view selection strategies on VGGT~\cite{wang2025vggt} with the ETH3D~\cite{schops2017eth3d} \textit{courtyard} scene. The top row shows the top-down view of the camera trajectory, with each view colored by its SO(3) bin index relative to the target view and the target view marked with a \textcolor{triangleyellow}{yellow} star; the bottom row shows the source view count per SO(3) bin. The two columns correspond to (a) an arbitrarily chosen target ($v = 0$) and (b) our proposed \textit{GRV} (Sec.~\ref{para:grv}) target ($v^{\ast} = 14$).}
\label{fig:ablation_fan}
\end{minipage}
\end{table*}

\begin{table*}[t]
\centering
\caption{Computational complexity of our proposed \textit{Self-Geometry} across six pretrained \textit{VFMs}. \textit{Add.\ Params.\ (M)}: \textit{LoRA} (Sec.~\ref{para:lora_adapter}) overhead (millions; \% of \textit{VFM} in parens). \textit{ETH3D per-scene adapt.\ time}: mean adaptation time on ETH3D, split into \textit{Init.} (LightGlue~\cite{lindenberger2023lightglue} matching + filtering), \textit{FAN} (Sec.~\ref{subsec:fan}), \textit{GDO} (50-iter \textit{TTA}; Sec.~\ref{subsec:gdo}), and \textit{Total}. Rightmost: mean \textit{Total} on 7Scenes~\cite{shotton2013scenes}, ScanNet++~\cite{yeshwanth2023scannetpp}, HiRoom~\cite{lin2025da3}. Measured on a single NVIDIA RTX PRO 6000 GPU~\cite{nvidia_rtxpro6000}.}
\label{tab:complexity}
\resizebox{0.8\textwidth}{!}{
\begin{tabular}{l|c|cccc|ccc}
\hline
\multirow{2}{*}{Model} &
\multirow{2}{*}{\shortstack{Add.\ Params.\ (M) $\downarrow$ \\ (vs.\ pretrained)}} &
\multicolumn{4}{c|}{ETH3D per-scene adapt. time (min.) $\downarrow$} &
\multicolumn{3}{c}{Total (min.) on other datasets $\downarrow$} \\
\cline{3-9}
 & & Init. & FAN & GDO & Total & 7Scenes & ScanNet++ & HiRoom \\
\hline
VGGT      & 18.87 (\textcolor{red}{\textbf{+1.5\%}}) & 0.30 & 0.01 & 2.98 & 3.29 & 5.36 & 5.83 & 2.22 \\
$\pi^3$   &  6.29 (\textcolor{red}{\textbf{+0.7\%}}) & 0.31 & 0.02 & 1.61 & 1.93 & 3.91 & 3.97 & 1.25 \\
DA3-Giant & 15.73 (\textcolor{red}{\textbf{+1.2\%}}) & 0.26 & 0.01 & 1.37 & 1.65 & 3.99 & 3.94 & 1.84 \\
DA3-Large &  6.29 (\textcolor{red}{\textbf{+1.5\%}}) & 0.26 & 0.01 & 0.98 & 1.25 & 3.39 & 3.31 & 1.42 \\
DA3-Base  &  2.36 (\textcolor{red}{\textbf{+1.7\%}}) & 0.26 & 0.02 & 0.39 & 0.66 & 2.68 & 2.47 & 0.61 \\
DA3-Small &  1.18 (\textcolor{red}{\textbf{+3.4\%}}) & 0.26 & 0.02 & 0.21 & 0.49 & 2.46 & 2.19 & 0.33 \\
\hline
\end{tabular}}
\end{table*}

\textbf{Ablations for FAN.}\label{para:ablation_fan} We ablate the proposed \textit{FAN} (Sec.~\ref{subsec:fan}) in Tab.~\ref{tab:ablation_fan} and analyze it in Fig.~\ref{fig:ablation_fan}. 15$^\circ$ (Tab.~\ref{tab:ablation_fan}-(i)-A, Ours), 30$^\circ$ (Tab.~\ref{tab:ablation_fan}-(i)-B), and 45$^\circ$ (Tab.~\ref{tab:ablation_fan}-(i)-C) all attain comparable performance across the three SO(3) bin widths, showing that the proposed \textit{FAN} is robust to the bin width hyperparameter. \textit{GRV} (Tab.~\ref{tab:ablation_fan}-(ii)-B, Ours), the target view that the proposed \textit{FAN} selects once using the baseline model prediction, improves substantially over Fixed $t=0$ (Tab.~\ref{tab:ablation_fan}-(ii)-A) on every column, confirming that target view selection is the dominant factor. In Fig.~\ref{fig:ablation_fan}, the SO(3) bin distributions of (a) Fixed $t=0$ (Tab.~\ref{tab:ablation_fan}-(ii)-A) and (b) \textit{GRV} (Tab.~\ref{tab:ablation_fan}-(ii)-B, Ours) visually show how each target view selection strategy covers the target scene. Dynamic \textit{GRV} (Tab.~\ref{tab:ablation_fan}-(ii)-C), which updates the target view every TTA iteration from the adapting model, underperforms \textit{GRV} (Tab.~\ref{tab:ablation_fan}-(ii)-B, Ours) since the adapting model is still evolving. Sequential (Tab.~\ref{tab:ablation_fan}-(iii)-A, Ours), Asc.\ Match Count (Tab.~\ref{tab:ablation_fan}-(iii)-B), and Desc.\ Match Count (Tab.~\ref{tab:ablation_fan}-(iii)-C) all attain comparable performance across all three orderings, confirming that the proposed \textit{FAN} is a robust view sampler.

\subsection{Complexity Comparison}
\label{subsec:complexity}

We further profile the computational complexity of our proposed \textit{Self-Geometry} across six pretrained \textit{VFMs} in Tab.~\ref{tab:complexity}. The \textit{LoRA} adapter adds only 0.7\%--3.4\% additional parameters over the pretrained \textit{VFM}, confirming that the proposed \textit{Lightweight TTA} strategy is parameter-efficient. Decomposing the per-scene adaptation wall-clock, \textit{Init.} takes about 0.3 minutes independent of model size, the proposed \textit{FAN} takes only 0.01--0.02 minutes and is negligible, and the proposed \textit{GDO} (VGGT~\cite{wang2025vggt} 2.98 min down to DA3-Small~\cite{lin2025da3} 0.21 min) is dominated by the forward-backward computation cost of the \textit{VFM} itself. As a result, our proposed \textit{Self-Geometry} completes per-scene adaptation within two minutes (up to 40 input views, DA3-Giant on ETH3D~\cite{schops2017eth3d}) on a single NVIDIA RTX PRO 6000~\cite{nvidia_rtxpro6000}, and remains within a practical scene-wise adaptation budget on the other three datasets (7Scenes~\cite{shotton2013scenes}, ScanNet++~\cite{yeshwanth2023scannetpp}, HiRoom~\cite{lin2025da3}).

\section{Conclusion}
\label{sec:conclusion}

We proposed \textit{Self-Geometry}, a \textit{GT-free} and \textit{Plug-and-play} \textit{TTA} pipeline that mitigates the \textit{multi-view geometric inconsistency} of pretrained \textit{VFMs} (Fig.~\ref{fig:failure}). The key insight is that \textit{2D pixel correspondences} extracted at test-time by an external feature matcher (LightGlue~\cite{lindenberger2023lightglue}) can serve as \textit{pseudo GT} for imposing \textit{explicit multi-view geometric supervision} on pretrained \textit{VFMs}. Building on this insight, our proposed \textit{Self-Geometry} comprises three complementary components: \textit{GDO} (Sec.~\ref{subsec:gdo}), which disentangles the gradients of the \textit{point-to-point} \textit{MVC Loss} (Sec.~\ref{para:mvc}) and the \textit{point-to-line} \textit{EC Loss} (Sec.~\ref{para:ec}); \textit{FAN} (Sec.~\ref{subsec:fan}), an SO(3)-guided view sampler using \textit{scene-scale-invariant} SO(3) geodesic distances; and \textit{Lightweight TTA} (Sec.~\ref{subsec:lightweight_tta}), a \textit{LoRA}~\cite{hu2022lora}-based parameter-efficient per-scene adaptation strategy. Experiments across six pretrained \textit{VFMs} (VGGT~\cite{wang2025vggt}, $\pi^3$~\cite{yang2025pi3}, DA3-G/L/B/S~\cite{lin2025da3}) and four benchmark datasets (7Scenes~\cite{shotton2013scenes}, ETH3D~\cite{schops2017eth3d}, ScanNet++~\cite{yeshwanth2023scannetpp}, HiRoom~\cite{lin2025da3}) confirm consistent improvements in both Pose Estimation and Geometry Estimation (Tab.~\ref{tab:pose_main}, Tab.~\ref{tab:geometry_main}, Fig.~\ref{fig:qual}). By completing per-scene adaptation within two minutes (up to 40 input views, DA3-Giant on ETH3D~\cite{schops2017eth3d}) on a single NVIDIA RTX PRO 6000~\cite{nvidia_rtxpro6000} GPU (Tab.~\ref{tab:complexity}), our proposed \textit{Self-Geometry} offers a practical recipe for augmenting frozen pretrained \textit{VFMs} with \textit{explicit multi-view geometric consistency} at test-time.

\bibliographystyle{IEEEtran}
\bibliography{arXiv/references}


\ifdefined\ARXIVMAINDOC
  \clearpage
  \setcounter{section}{0}
  \setcounter{subsection}{0}
  \setcounter{equation}{0}
  \setcounter{figure}{0}
  \setcounter{table}{0}
  \renewcommand{\thesection}{S.\Roman{section}}
  \renewcommand{\thesubsection}{\Alph{subsection}}
  \renewcommand{\theequation}{S.\arabic{equation}}
  \renewcommand{\thefigure}{S.\arabic{figure}}
  \renewcommand{\thetable}{S.\arabic{table}}

  \begin{center}\Large\textbf{Supplementary Materials}\end{center}
  \vspace{0.5em}


\section{GitHub Repository}
\label{sec:supp_github}

The code will be released soon at \url{https://github.com/CMLab-Korea/Self-Geometry}.

\section{Preliminary: Epipolar Geometry}
\label{sec:preliminary}

In this section, we summarize the key concepts of epipolar geometry used in Sec.~\ref{sec:method}. Epipolar geometry defines the geometric relation that correspondences between two views must satisfy, and forms the basis of our \textit{Epipolar Consistency Loss} (\textit{EC Loss}) (Eq.~\eqref{eq:ec}). To this end, we consider two views $(I_i, I_j)$, a target view $I_i$ and a source view $I_j$, where $I_i, I_j \in \mathbb{R}^{H \times W \times 3}$ and $H$ and $W$ are the vertical and horizontal image resolutions, respectively. The camera intrinsic matrix of each view is denoted by $\mathbf{K}_i, \mathbf{K}_j \in \mathbb{R}^{3 \times 3}$, respectively, and the relative pose from the source view $j$ to the target view $i$ is denoted by $[\mathbf{R}_{i \leftarrow j} \mid \mathbf{t}_{i \leftarrow j}] \in \mathbb{R}^{3 \times 4}$, where $\mathbf{R}_{i \leftarrow j} \in SO(3)$ is a rotation matrix and $\mathbf{t}_{i \leftarrow j} \in \mathbb{R}^{3}$ is a translation vector. The correspondence pair between the two views is represented by the homogeneous 2D coordinates $\tilde{\mathbf{x}}_i, \tilde{\mathbf{x}}_j \in \mathbb{R}^{3}$ in the pixel coordinate system.

Under the above notation, if the target view and the source view observe the same 3D point, their $\tilde{\mathbf{x}}_i, \tilde{\mathbf{x}}_j$ must satisfy a linear constraint derived from $[\mathbf{R}_{i \leftarrow j} \mid \mathbf{t}_{i \leftarrow j}]$. We refer to this constraint as the \textit{epipolar relation}, which is most concisely expressed in the calibrated normalized coordinate system where the effect of $\mathbf{K}_i, \mathbf{K}_j$ is removed. This expression is encoded by the \textit{essential matrix} defined from $[\mathbf{R}_{i \leftarrow j} \mid \mathbf{t}_{i \leftarrow j}]$:
\begin{equation}
\label{eq:essential}
\mathbf{E}_{i \leftarrow j} = [\mathbf{t}_{i \leftarrow j}]_{\times} \mathbf{R}_{i \leftarrow j},
\end{equation}
where $[\mathbf{t}_{i \leftarrow j}]_{\times} \in \mathbb{R}^{3 \times 3}$ is the skew-symmetric matrix of $\mathbf{t}_{i \leftarrow j}$. In contrast, since $\tilde{\mathbf{x}}_i, \tilde{\mathbf{x}}_j$ are defined in the pixel coordinate system, the epipolar relation holding in the calibrated normalized coordinate system must be directly expressed in the pixel coordinate system. To this end, we define the \textit{fundamental matrix} as follows:
\begin{equation}
\label{eq:fundamental}
\mathbf{F}_{i \leftarrow j} = \mathbf{K}_i^{-\top} \mathbf{E}_{i \leftarrow j}\, \mathbf{K}_j^{-1}.
\end{equation}

Here, $\mathbf{F}_{i \leftarrow j}$ maps a correspondence in one view to an \textit{epipolar line} in the other view. Specifically, given $\tilde{\mathbf{x}}_j$ in the source view, the epipolar line $\boldsymbol{\ell}_i \in \mathbb{R}^{3}$ on which the corresponding correspondence in the target view must lie is defined as:
\begin{equation}
\label{eq:epi_line}
\boldsymbol{\ell}_i = \mathbf{F}_{i \leftarrow j}\, \tilde{\mathbf{x}}_j.
\end{equation}
If the target view and the source view observe the same 3D point, $\tilde{\mathbf{x}}_i$ in the target view must lie on $\boldsymbol{\ell}_i$ (Eq.~\eqref{eq:epi_line}). This relation is algebraically expressed as the \textit{epipolar constraint}:
\begin{equation}
\label{eq:epi_constraint}
\tilde{\mathbf{x}}_i^{\top} \mathbf{F}_{i \leftarrow j}\, \tilde{\mathbf{x}}_j = 0.
\end{equation}
Therefore, the epipolar constraint (Eq.~\eqref{eq:epi_constraint}) provides the basic condition for determining whether a correspondence pair and the relative camera pose are geometrically consistent.

In practice, since $\mathbf{F}_{i \leftarrow j}$ and correspondences contain noise, it is difficult to exactly satisfy the epipolar constraint (Eq.~\eqref{eq:epi_constraint}). This requires a residual that quantifies the degree of violation of the epipolar constraint, for which we use the \textit{Sampson distance}~\cite{hartley2003multiple}, defined as follows:
\begin{equation}
\label{eq:sampson}
\resizebox{0.85\hsize}{!}{$
d_{\mathrm{Sampson}}\!\left(\tilde{\mathbf{x}}_i, \tilde{\mathbf{x}}_j;\, \mathbf{F}_{i \leftarrow j}\right) = \dfrac{ (\tilde{\mathbf{x}}_i^{\top} \mathbf{F}_{i \leftarrow j}\, \tilde{\mathbf{x}}_j)^{2} }{ \sum_{k=1}^{2} \left[ (\mathbf{F}_{i \leftarrow j}\, \tilde{\mathbf{x}}_j)_{k}^{2} + (\mathbf{F}_{i \leftarrow j}^{\top} \tilde{\mathbf{x}}_i)_{k}^{2} \right] }
$}.
\end{equation}
Here, $(\cdot)_k$ denotes the $k$-th component of a vector. Since the Sampson distance robustly quantifies how much a correspondence pair violates the epipolar constraint, it is directly used as the residual in the \textit{EC Loss} (Eq.~\eqref{eq:ec}) in Sec.~\ref{subsec:gdo}.

\section{Total Loss Formulation}
\label{sec:supp_total_loss}

In this section, we formalize the total loss that our proposed \textit{Self-Geometry} optimizes at each \textit{Test-Time Adaptation (TTA)} iteration $t$. The total loss combines the two primary losses introduced in Sec.~\ref{sec:method} ($\mathcal{L}_{\mathrm{mvc}}$, Eq.~\eqref{eq:mvc}; $\mathcal{L}_{\mathrm{ec}}$, Eq.~\eqref{eq:ec}) with three self-supervised auxiliary regularizers ($\mathcal{L}_{\mathrm{pc}}$, $\mathcal{L}_{\mathrm{eds}}$, $\mathcal{L}_{\mathrm{bdc}}$) that suppress excessive drift of the pretrained \textit{VFM} depth predictions from the target scene. The gradients of the two primary losses are disentangled at every \textit{TTA} iteration via our proposed \textit{GD} (Eq.~\eqref{eq:gdo_projection}), and every reprojection or distance based residual passes through \textit{Huber Robustification} (Eq.~\eqref{eq:huber_loss}) to guard against outlier-induced instability.

\textbf{Photometric Consistency Loss, $\mathcal{L}_{\mathrm{pc}}$.} Following Monodepth2~\cite{godard2019monodepth2}, this loss measures the photometric error between the target view $\mathbf{I}_i$ and the image $\hat{\mathbf{I}}_{i \leftarrow j}$ obtained by warping source view $j$ into target view $i$ using the camera poses $\mathcal{P}$ and depth $\mathbf{D}_i$ predicted by the pretrained \textit{VFM}. This loss mixes SSIM and L1 as:
\begin{equation}
\label{eq:pc}
\mathcal{L}_{\mathrm{pc}} = \alpha \cdot \frac{1 - \mathrm{SSIM}(\mathbf{I}_i, \hat{\mathbf{I}}_{i \leftarrow j})}{2} + (1 - \alpha)\bigl\|\mathbf{I}_i - \hat{\mathbf{I}}_{i \leftarrow j}\bigr\|_1.
\end{equation}
Here, SSIM denotes the structural similarity index~\cite{wang2004ssim} between two images, and $\alpha \in [0, 1]$ is the mixing weight ($\alpha = 0.85$).

\textbf{Edge-aware Depth Smoothness Loss, $\mathcal{L}_{\mathrm{eds}}$.} Following Monodepth2~\cite{godard2019monodepth2}, this loss encourages the depth map to remain discontinuous along image edges and smooth elsewhere. It is computed as the L1 norm of the depth gradient exponentially weighted by the image gradient:
\begin{equation}
\label{eq:eds}
\mathcal{L}_{\mathrm{eds}} = \bigl|\partial_x \mathbf{D}_i\bigr| \, e^{-|\partial_x \mathbf{I}_i|} + \bigl|\partial_y \mathbf{D}_i\bigr| \, e^{-|\partial_y \mathbf{I}_i|}.
\end{equation}
Here, $\partial_x$ and $\partial_y$ denote the horizontal and vertical spatial gradient operators, respectively.

\textbf{Baseline Depth Consistency Loss, $\mathcal{L}_{\mathrm{bdc}}$.} This loss uses the baseline depth prediction $\mathbf{D}_i^{\mathrm{base}}$ of the pretrained \textit{VFM} as an anchor to suppress excessive drift of the adapted depth $\mathbf{D}_i$. It is given by the L1 error restricted to the confident-pixel set $\mathcal{V}_c$ that comprises the top $q$-quantile of the baseline confidence:
\begin{equation}
\label{eq:bdc}
\mathcal{L}_{\mathrm{bdc}} = \frac{1}{|\mathcal{V}_c|} \sum_{p \in \mathcal{V}_c} \bigl| \mathbf{D}_i(p) - \mathbf{D}_i^{\mathrm{base}}(p) \bigr|.
\end{equation}
Here, $q = 0.5$ is the quantile that controls the size of the confident region.

\textbf{Huber Robustification.} The per-point residuals of $\mathcal{L}_{\mathrm{mvc}}, \mathcal{L}_{\mathrm{ec}}, \mathcal{L}_{\mathrm{pc}}$, and $\mathcal{L}_{\mathrm{bdc}}$ pass through classical Huber loss to guard against outlier-induced instability ($\mathcal{L}_{\mathrm{eds}}$ is used unmodified since it is already spatially regularized). The threshold $\delta$ is set from a median absolute deviation (MAD) based scale $\sigma$~\cite{szeliski2022computer}. The robust scale $\sigma$ and the Huber threshold $\delta$ are respectively given by:
\begin{equation}
\label{eq:huber_delta}
\sigma = 1.4826 \cdot \mathrm{median}(|r|), \quad \delta = 1.345 \cdot \sigma,
\end{equation}
and the Huber loss $H_\delta(r)$ takes the standard piecewise form:
\begin{equation}
\label{eq:huber_loss}
H_\delta(r) = \begin{cases} \tfrac{1}{2} r^2 & \text{if } |r| \le \delta, \\ \delta \bigl(|r| - \tfrac{1}{2}\delta\bigr) & \text{otherwise}. \end{cases}
\end{equation}
Here, $r$ denotes the per-point residual of each loss. The $\delta$ of the primary losses is set once at scene initialization from the baseline forward residual (after removing the top 10\% outliers) and remains fixed across iterations, whereas the $\delta$ of the auxiliary $\mathcal{L}_{\mathrm{bdc}}$ is re-estimated at each iteration from the residual between the adapted forward and the baseline.

\textbf{Dynamic Weight Averaging.} To prevent optimizer bias toward any single loss caused by scale imbalance across losses, we apply \textit{Dynamic Weight Averaging (DWA)}~\cite{liu2019dwa} to dynamically rebalance the weight $w_k(t)$ of each loss at every iteration. The recent two-iteration decay ratio $r_k(t)$ of each loss $k$ and its softmax-normalized weight $w_k(t)$ are computed as:
\begin{equation}
\label{eq:dwa}
r_k(t) = \frac{\mathcal{L}_k(t-1)}{\mathcal{L}_k(t-2)}, \quad w_k(t) = K \cdot \frac{\exp\bigl(r_k(t)/T\bigr)}{\sum_{j=1}^{K} \exp\bigl(r_j(t)/T\bigr)}.
\end{equation}
Here, $K = 5$ is the number of losses that constitute the total loss, and $T = 1.0$ is the softmax temperature. To prevent extreme rebalancing, $r_k(t)$ is clamped to $[0.5, 2.0]$.

\textbf{Total Loss.} Combining the primary and auxiliary losses, the total loss is expressed as the sum of the Huber-robustified form $\tilde{\mathcal{L}}_k$ of each loss weighted by both a static weight $\lambda_k$ and the dynamic DWA weight $w_k(t)$:
\begin{equation}
\label{eq:total_loss}
\mathcal{L}_{\mathrm{total}}(t) = \sum_{k \in \{\mathrm{mvc},\, \mathrm{ec},\, \mathrm{pc},\, \mathrm{eds},\, \mathrm{bdc}\}} w_k(t) \, \lambda_k \, \tilde{\mathcal{L}}_k.
\end{equation}
Here, $\tilde{\mathcal{L}}_{\mathrm{eds}} = \mathcal{L}_{\mathrm{eds}}$ (unmodified), and every other $\tilde{\mathcal{L}}_k$ applies the Huber loss $H_\delta$ (Eq.~\eqref{eq:huber_loss}) to its per-point residuals. Our default configuration sets $\lambda_k = 1.0$ for every loss. At each gradient step, $\nabla \tilde{\mathcal{L}}_{\mathrm{mvc}}$ is projected onto the orthogonal complement of $\nabla \tilde{\mathcal{L}}_{\mathrm{ec}}$ via our proposed \textit{GD} (Eq.~\eqref{eq:gdo_projection}) before summation into the total gradient.

\begin{algorithm*}[!t]
\caption{Per-scene TTA of our proposed \textit{Self-Geometry}.}
\label{alg:supp_selfgeometry}
\footnotesize
\begin{algorithmic}[1]
\Require Input views $\mathcal{I} = \{\mathbf{I}_i\}_{i=1}^{N}$; frozen pretrained \textit{Vision Foundation Model (VFM)} $f_\theta$; initialized \textit{Low-Rank Adaptation (LoRA)} parameters $\phi$; total iterations $T$; SO(3) bin width $\Delta\theta$; AdamW learning rate $\eta$; gradient-norm clip $C$.
\Ensure Adapted \textit{LoRA} parameters $\phi^{\ast}$.
\State Compute the baseline pose $\mathcal{P}^{\mathrm{base}}$ and depth $\mathcal{D}^{\mathrm{base}}$: $(\mathcal{P}^{\mathrm{base}}, \mathcal{D}^{\mathrm{base}}) \gets f_\theta(\mathcal{I})$.
\State $\mathcal{M} \gets$ LightGlue~\cite{lindenberger2023lightglue} matches on every view pair (Eq.~\eqref{eq:pseudo_corr_set}); apply \textit{Pseudo-Correspondence Filtering} with the $\mathcal{L}_{\mathrm{ec}}$-based filter followed by the $\mathcal{L}_{\mathrm{mvc}}$-based filter.
\State Estimate per-scene Huber thresholds $\delta_{\mathrm{mvc}}, \delta_{\mathrm{ec}}$ from baseline residuals on $\mathcal{M}$ (Eq.~\eqref{eq:huber_delta}).
\State Assign source views to SO(3) bins of width $\Delta\theta$ and select the target view $v^{\ast} \gets \arg\max_{v \in \mathcal{V}} H(v)$ via \textit{Geometry-Rich View Selection} (GRV; Eqs.~\eqref{eq:so3_angle}, \eqref{eq:grv_selection}).
\State Initialize the running best checkpoint metric $\mathcal{L}_{\mathrm{best}} \gets \infty$ and the running best \textit{LoRA} parameters $\phi_{\mathrm{best}} \gets \phi$.
\For{$t = 1, \dots, T$}
    \State Sample the source view subset $\mathcal{I}_t$ from bins around $v^{\ast}$ via \textit{Angular-Neighbor Sampling (ANS)}.
    \State Forward $\mathcal{I}_t$ through the \textit{LoRA}-adapted \textit{VFM} $f_{\theta, \phi}$ to obtain the adapted pose $\mathcal{P}_t$ and depth $\mathcal{D}_t$: $(\mathcal{P}_t, \mathcal{D}_t) \gets f_{\theta, \phi}(\mathcal{I}_t)$.
    \State Compute the primary losses $\mathcal{L}_{\mathrm{mvc}}, \mathcal{L}_{\mathrm{ec}}$ (Eqs.~\eqref{eq:mvc}, \eqref{eq:ec}) and the auxiliary regularizers $\mathcal{L}_{\mathrm{pc}}, \mathcal{L}_{\mathrm{eds}}, \mathcal{L}_{\mathrm{bdc}}$ (Eqs.~\eqref{eq:pc}--\eqref{eq:bdc}).
    \State Apply \textit{Huber Robustification} (Eq.~\eqref{eq:huber_loss}) to obtain $\tilde{\mathcal{L}}_k \gets H_\delta(\mathcal{L}_k)$ for $k \in \{\mathrm{mvc}, \mathrm{ec}, \mathrm{pc}, \mathrm{bdc}\}$; set $\tilde{\mathcal{L}}_{\mathrm{eds}} \gets \mathcal{L}_{\mathrm{eds}}$.
    \State Rebalance via \textit{Dynamic Weight Averaging}: update $w_k(t)$ (Eq.~\eqref{eq:dwa}) and form $\mathcal{L}_{\mathrm{total}}(t)$ (Eq.~\eqref{eq:total_loss}).
    \State Compute per-task gradients $\nabla \tilde{\mathcal{L}}_{\mathrm{mvc}}, \nabla \tilde{\mathcal{L}}_{\mathrm{ec}}$ with respect to $\phi$.
    \State Apply \textit{Gradient Disentanglement (GD)}: project $\nabla \tilde{\mathcal{L}}_{\mathrm{mvc}}$ onto the orthogonal complement of $\nabla \tilde{\mathcal{L}}_{\mathrm{ec}}$ (Eq.~\eqref{eq:gdo_projection}).
    \State Aggregate all gradients, clip the global norm to $C$, and update $\phi$ via AdamW$(\eta)$.
    \State Compute scale-invariant checkpoint metric $s(t) \gets \sqrt{\tilde{\mathcal{L}}_{\mathrm{ec}}(t) \cdot \tilde{\mathcal{L}}_{\mathrm{mvc}}(t)}$; \textbf{if} $s(t) < \mathcal{L}_{\mathrm{best}}$ \textbf{then} $\mathcal{L}_{\mathrm{best}} \gets s(t)$, $\phi_{\mathrm{best}} \gets \phi$.
\EndFor
\State \Return $\phi^{\ast} \gets \phi_{\mathrm{best}}$.
\end{algorithmic}
\normalsize
\end{algorithm*}

\section{Algorithm}
\label{sec:supp_algorithm}

Algorithm~\ref{alg:supp_selfgeometry} summarizes the complete per-scene procedure of our proposed \textit{Self-Geometry}. The procedure comprises three stages: (i) \textit{Scene Initialization}, which extracts the pseudo-correspondence set $\mathcal{M}$ via LightGlue~\cite{lindenberger2023lightglue}, applies \textit{Pseudo-Correspondence Filtering}, and estimates the per-scene Huber thresholds (Eq.~\eqref{eq:huber_delta}); (ii) target view selection via our proposed \textit{GRV}; and (iii) $T$ iterations of the \textit{TTA} loop, each performing our proposed \textit{ANS}, computing the primary losses ($\mathcal{L}_{\mathrm{mvc}}$, $\mathcal{L}_{\mathrm{ec}}$) together with the three auxiliary regularizers ($\mathcal{L}_{\mathrm{pc}}$, $\mathcal{L}_{\mathrm{eds}}$, $\mathcal{L}_{\mathrm{bdc}}$; Eqs.~\eqref{eq:pc}--\eqref{eq:bdc}), applying \textit{Huber Robustification} (Eq.~\eqref{eq:huber_loss}) and \textit{Dynamic Weight Averaging} (Eq.~\eqref{eq:dwa}), and taking an AdamW step after our proposed \textit{GD} on the primary loss gradients.

\begin{table}
\centering
\caption{TTA hyperparameters of our proposed \textit{Self-Geometry}. Every value in this table is used identically across all combinations of the six pretrained \textit{VFMs} and four benchmark datasets, without any per-\textit{VFM} or per-dataset tuning.}
\label{tab:supp_hparams}
\begin{tabular}{l|l}
\hline
\multicolumn{2}{l}{Hyperparameter} \\
\hline
Iterations $T$ & $50$ \\
\hline
\multicolumn{2}{l}{\textit{Optimizer}} \\
\hline
Type & AdamW~\cite{loshchilov2019adamw} \\
Learning rate $\eta$ & $5 \times 10^{-5}$ \\
Weight decay & $0.05$ \\
Schedule & cosine with $5\%$ warmup \\
Final learning rate & $1 \times 10^{-8}$ \\
Gradient clip $C$ & $5.0$ \\
\hline
\multicolumn{2}{l}{\textit{LoRA} (Sec.~\ref{subsec:lightweight_tta})} \\
\hline
Rank $r$ & $64$ \\
Alpha $\alpha_L$ & $64$ \\
Dropout & $0.1$ \\
\hline
\end{tabular}
\end{table}

\section{Training Details}
\label{sec:supp_training}

Table~\ref{tab:supp_hparams} lists the hyperparameters used by our proposed \textit{Self-Geometry}. \textbf{Every value in Table~\ref{tab:supp_hparams} is used identically across all combinations of the six pretrained \textit{VFMs} (VGGT~\cite{wang2025vggt}, $\pi^3$~\cite{yang2025pi3}, DA3-Giant/Large/Base/Small~\cite{lin2025da3}) and the four benchmark datasets (7Scenes~\cite{shotton2013scenes}, ETH3D~\cite{schops2017eth3d}, ScanNet++~\cite{yeshwanth2023scannetpp}, HiRoom~\cite{lin2025da3}) in our experiments.} Following Sec.~\ref{subsec:lightweight_tta}, we insert \textit{LoRA}~\cite{hu2022lora} into the QKV weights of every attention block of each pretrained \textit{VFM}, keeping all remaining parameters of the pretrained \textit{VFM} frozen and updating only the \textit{LoRA} parameters. This \textit{LoRA} adapter configuration is applied identically across all six pretrained \textit{VFMs}, except that for $\pi^3$~\cite{yang2025pi3} \textit{LoRA} is inserted only into the encoder due to its model size.

\begin{figure*}[!t]
\centering
\includegraphics[height=8cm,keepaspectratio]{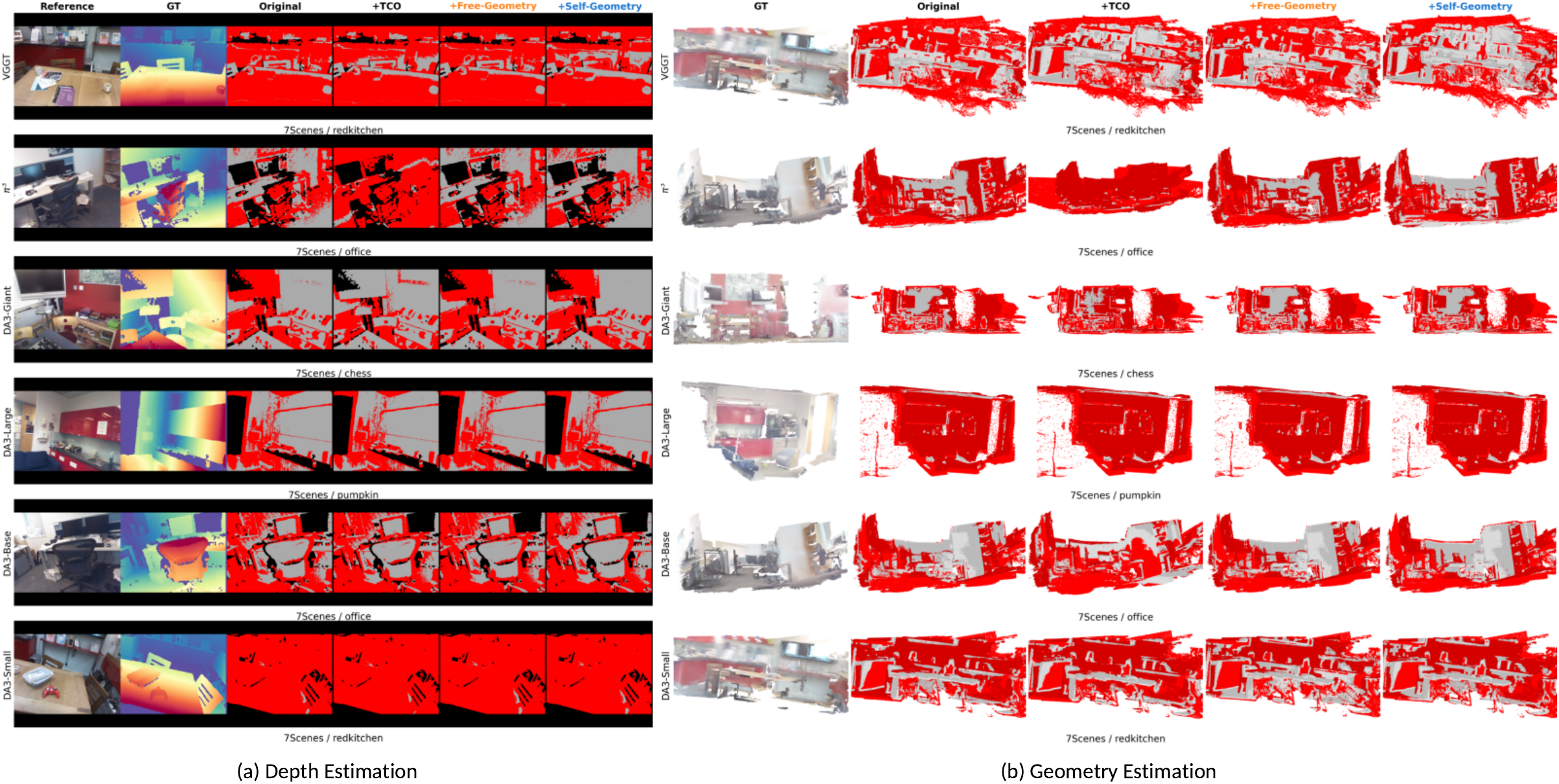}
\caption{Additional qualitative results on 7Scenes~\cite{shotton2013scenes}: rows are six pretrained \textit{VFMs}, each showing the scene with the largest F1 improvement for that \textit{VFM}. \textbf{(a) Depth Estimation}: from left to right, Reference RGB, GT depth, and depth error maps. \textbf{(b) Geometry Estimation}: from left to right, GT fused pointcloud and geometry error maps. Both (a) and (b) show Original, +TCO~\cite{tco}, \textcolor{orange}{+Free-Geometry~\cite{dai2026freegeometry}}, and \textcolor{blue}{+Self-Geometry} (Ours), where \textcolor{red}{red}/\textcolor{gray}{gray} denote errors above/within the benchmark threshold. \textit{Best viewed in zoom.}}
\label{fig:supp_qual_7scenes}
\end{figure*}
\begin{figure*}[!t]
\centering
\includegraphics[height=8cm,keepaspectratio]{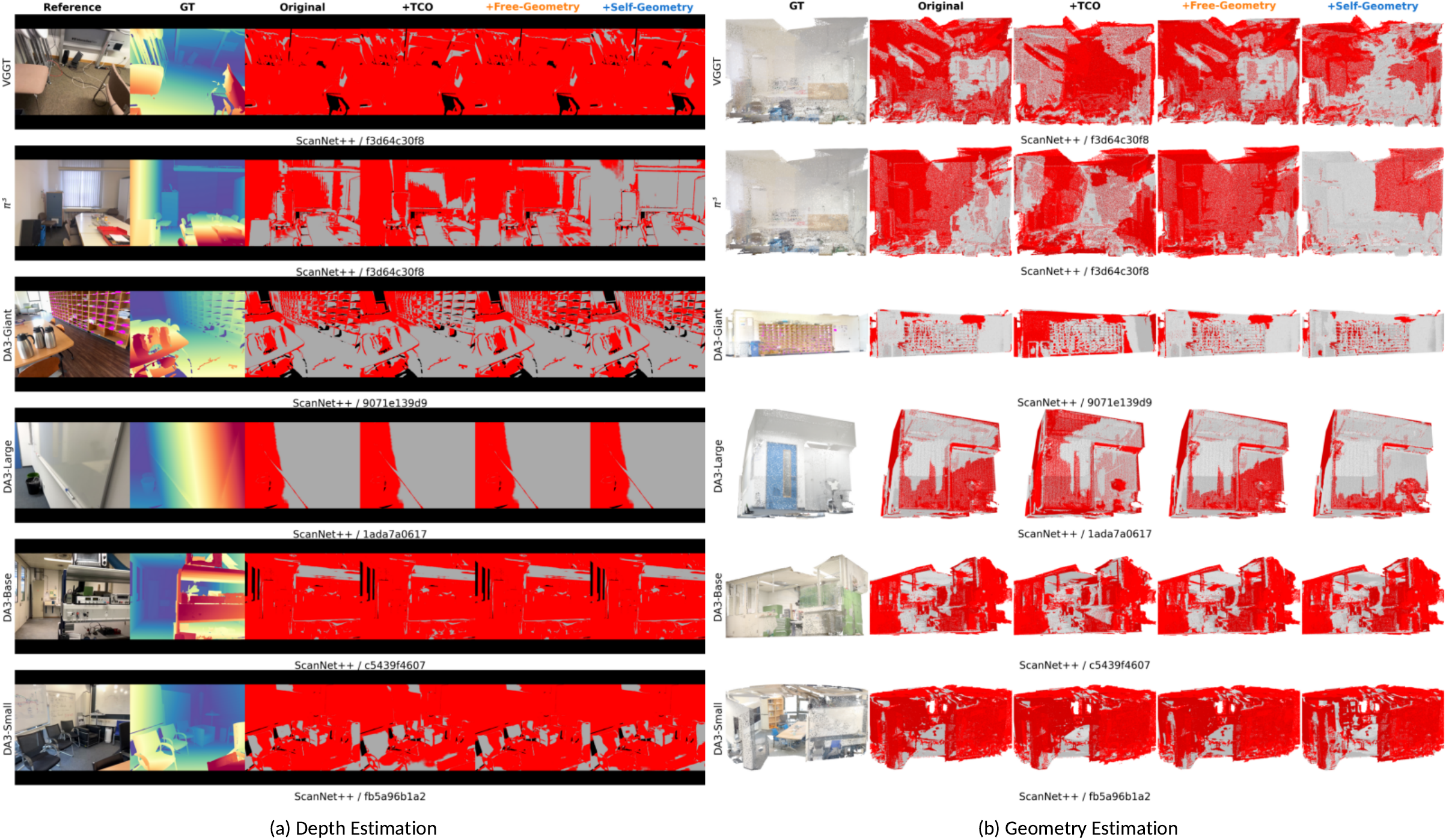}
\caption{Additional qualitative results on ScanNet++~\cite{yeshwanth2023scannetpp}: rows are six pretrained \textit{VFMs}, each showing the scene with the largest F1 improvement for that \textit{VFM}. \textbf{(a) Depth Estimation}: from left to right, Reference RGB, GT depth, and depth error maps. \textbf{(b) Geometry Estimation}: from left to right, GT fused pointcloud and geometry error maps. Both (a) and (b) show Original, +TCO~\cite{tco}, \textcolor{orange}{+Free-Geometry~\cite{dai2026freegeometry}}, and \textcolor{blue}{+Self-Geometry} (Ours), where \textcolor{red}{red}/\textcolor{gray}{gray} denote errors above/within the benchmark threshold. \textit{Best viewed in zoom.}}
\label{fig:supp_qual_scannetpp}
\end{figure*}
\begin{figure*}[!t]
\centering
\includegraphics[height=8cm,keepaspectratio]{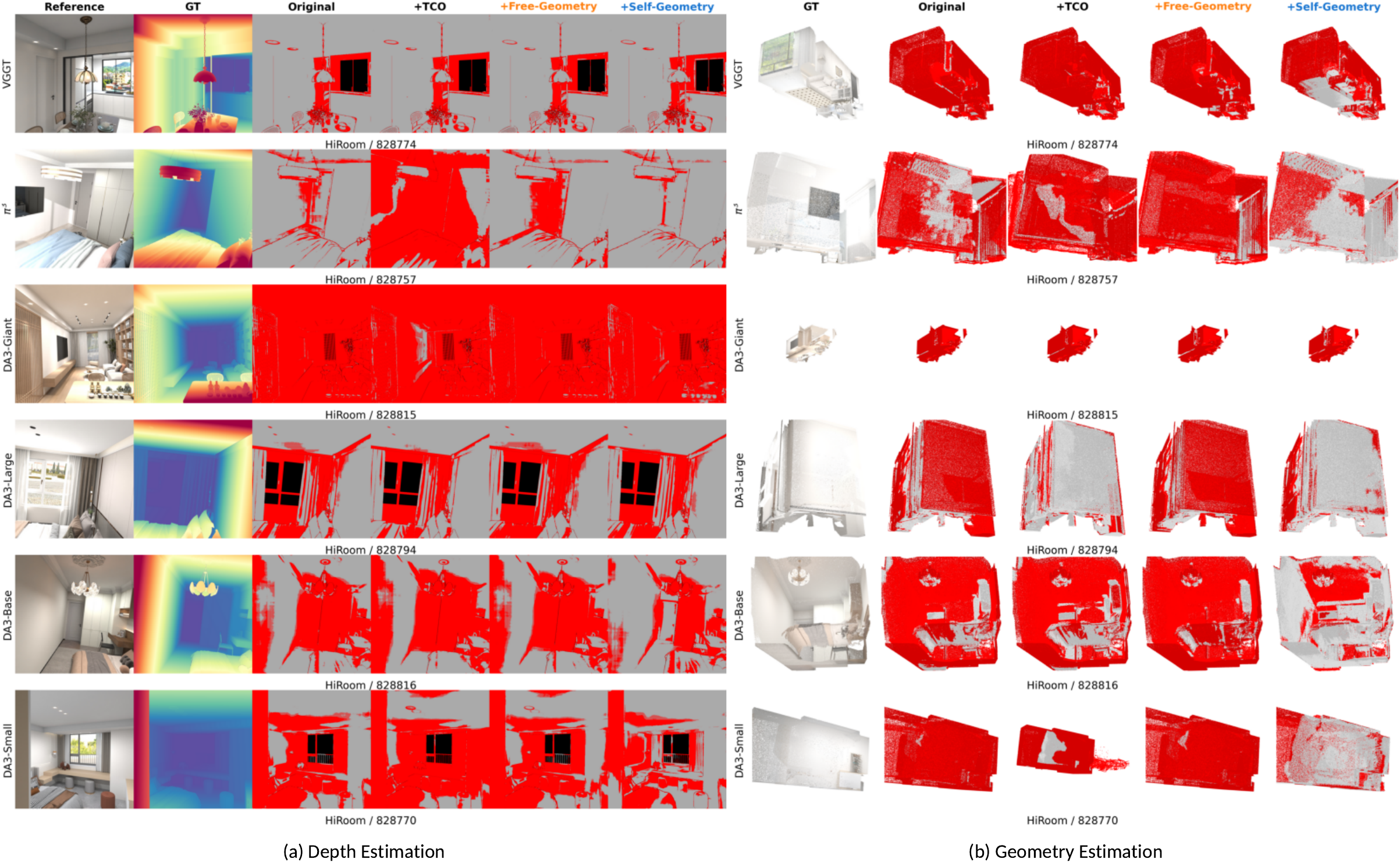}
\caption{Additional qualitative results on HiRoom~\cite{lin2025da3}: rows are six pretrained \textit{VFMs}, each showing the scene with the largest F1 improvement for that \textit{VFM}. \textbf{(a) Depth Estimation}: from left to right, Reference RGB, GT depth, and depth error maps. \textbf{(b) Geometry Estimation}: from left to right, GT fused pointcloud and geometry error maps. Both (a) and (b) show Original, +TCO~\cite{tco}, \textcolor{orange}{+Free-Geometry~\cite{dai2026freegeometry}}, and \textcolor{blue}{+Self-Geometry} (Ours), where \textcolor{red}{red}/\textcolor{gray}{gray} denote errors above/within the benchmark threshold. \textit{Best viewed in zoom.}}
\label{fig:supp_qual_hiroom}
\end{figure*}
\section{Limitations}
\label{sec:supp_limitations}

\textbf{Dependence on External Feature Matcher.} Our proposed \textit{Self-Geometry} obtains its multi-view geometric supervision from the 2D pixel correspondences produced by an external feature matcher (LightGlue~\cite{lindenberger2023lightglue}). Consequently, on scenes where the matcher struggles (\textit{e.g.}, repetitive textures, textureless surfaces, or wide-baseline viewpoint changes with limited overlap), the supervision signal weakens and adaptation quality degrades.

\textbf{Adaptation Latency.} Although our proposed \textit{Self-Geometry} completes per-scene adaptation within a few minutes on a single GPU (Tab.~\ref{tab:complexity}), this latency remains far from real-time requirements.

\else
  \documentclass[10pt, journal, twoside]{IEEEtran}

  \usepackage[utf8]{inputenc}
  \usepackage[T1]{fontenc}
  \usepackage{amsmath, amssymb, amsfonts}
  \usepackage{mathtools}
  \usepackage{algorithm}
  \usepackage{algpseudocode}
  \usepackage{graphicx}
  \usepackage{booktabs}
  \usepackage{multirow}
  \usepackage{array}
  \usepackage{xcolor}
  \usepackage{cite}
  \usepackage{url}
  \usepackage[hidelinks]{hyperref}

  \renewcommand{\thesection}{S.\Roman{section}}
  \renewcommand{\thesubsection}{\Alph{subsection}}
  \renewcommand{\theequation}{S.\arabic{equation}}
  \renewcommand{\thefigure}{S.\arabic{figure}}
  \renewcommand{\thetable}{S.\arabic{table}}

  \begin{document}

  \title{Supplementary Materials to ``Self-Geometry: GT-Free and Plug-and-Play Test-Time Adaptation for Geometrically Consistent 3D Vision Foundation Models''}
  \author{}

  \maketitle

  \bibliographystyle{IEEEtran}
  \bibliography{arXiv/references}

  \end{document}
\fi

\end{document}